\documentclass[sigconf]{acmart}

\AtBeginDocument{%
  \providecommand\BibTeX{{%
    \normalfont B\kern-0.5em{\scshape i\kern-0.25em b}\kern-0.8em\TeX}}}

\usepackage{subcaption}
\usepackage{diagbox}
\usepackage{multirow}
\usepackage{multicol}
\usepackage{algorithm}
\usepackage{algorithmic}

\copyrightyear{2022}
\acmYear{2022}
\setcopyright{acmlicensed}\acmConference[CIKM '22]{Proceedings of the 31st ACM International Conference on Information and Knowledge Management}{October 17--21, 2022}{Atlanta, GA, USA}
\acmBooktitle{Proceedings of the 31st ACM International Conference on Information and Knowledge Management (CIKM '22), October 17--21, 2022, Atlanta, GA, USA}
\acmPrice{15.00}
\acmDOI{10.1145/3511808.3557474}
\acmISBN{978-1-4503-9236-5/22/10}



\begin{document}

\title[Towards Automated Imbalanced Learning with Deep Hierarchical Reinforcement Learning]{Towards Automated Imbalanced Learning with Deep \\ Hierarchical Reinforcement Learning}

\author{Daochen Zha}
\author{Kwei-Herng Lai}
\email{daochen.zha@rice.edu}
\affiliation{%
  \institution{Rice University}
  \city{Houston}
  \state{TX}
  \country{USA}
}

\author{Qiaoyu Tan}
\author{Sirui Ding}
\author{Na Zou}
\affiliation{%
  \institution{Texas A\&M University}
  \city{College Station}
  \state{TX}
  \country{USA}
}

\author{Xia Hu}
\email{xia.hu@rice.edu}
\affiliation{%
  \institution{Rice University}
  \city{Houston}
  \state{TX}
  \country{USA}
}

\renewcommand{\shortauthors}{Daochen Zha et al.}

\begin{abstract}
Imbalanced learning is a fundamental challenge in data mining, where there is a disproportionate ratio of training samples in each class. Over-sampling is an effective technique to tackle imbalanced learning through generating synthetic samples for the minority class. While numerous over-sampling algorithms have been proposed, they heavily rely on heuristics, which could be sub-optimal since we may need different sampling strategies for different datasets and base classifiers, and they cannot directly optimize the performance metric. Motivated by this, we investigate  developing a learning-based over-sampling algorithm to optimize the classification performance, which is a challenging task because of the huge and hierarchical decision space. At the high level, we need to decide how many synthetic samples to generate. At the low level, we need to determine where the synthetic samples should be located, which depends on the high-level decision since the optimal locations of the samples may differ for different numbers of samples. To address the challenges, we propose AutoSMOTE, an automated over-sampling algorithm that can jointly optimize different levels of decisions. Motivated by the success of SMOTE~\cite{chawla2002smote} and its extensions, we formulate the generation process as a Markov decision process (MDP) consisting of three levels of policies to generate synthetic samples within the SMOTE search space. Then we leverage deep hierarchical reinforcement learning to optimize the performance metric on the validation data. Extensive experiments on six real-world datasets demonstrate that AutoSMOTE significantly outperforms the state-of-the-art resampling algorithms. The code is at \url{https://github.com/daochenzha/autosmote}

\end{abstract}


\maketitle

\section{Introduction}

Imbalanced learning is a fundamental challenge in many real-world applications, such as fraud detection, fake news detection, and medical diagnosis~\cite{shu2017fake,rout2018handling,johnson2019survey}, where there is a disproportionate ratio of training samples in each class. This phenomenon will negatively affect the classification performance since the standard classifiers will tend to be dominated by the majority class and perform poorly on the minority class~\cite{chawla2004special}. A common strategy to tackle imbalanced learning is resampling, which focuses on modifying the training data to balance the data distribution. In contrast to the algorithm-level solutions that modify the classifier~\cite{krawczyk2016learning}, resampling is argued to be more flexible as it does not make any assumption on the classifier so that it is generally applicable to various classifiers~\cite{santoso2017synthetic}.


\begin{figure}
    \centering
    \includegraphics[width=0.45\textwidth]{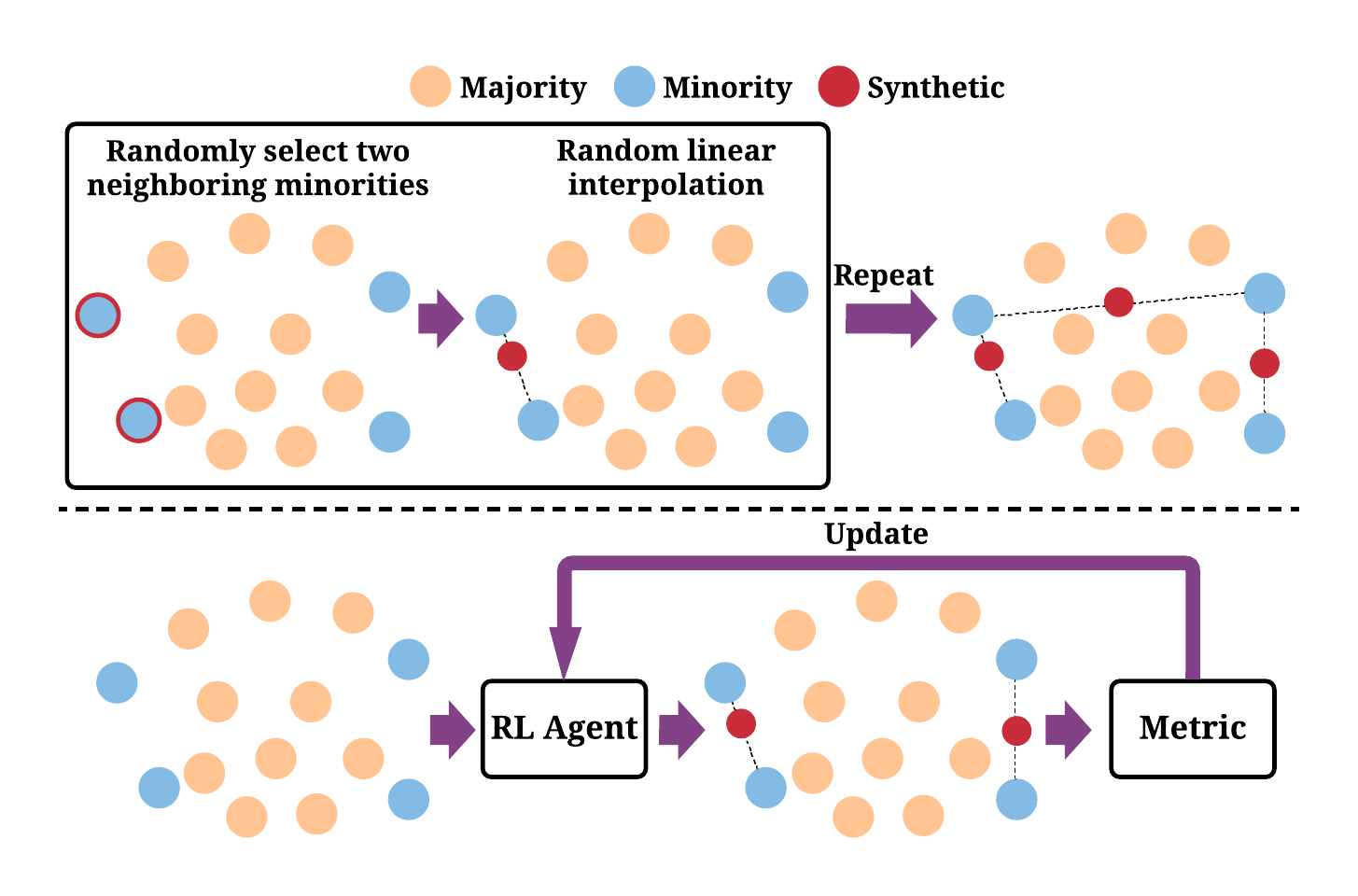}
    \vspace{-14pt}
    \caption{The decisions are randomly made in SMOTE (top) while the decisions in AutoSMOTE are made with an RL agent to optimize the performance on the validation set (bottom). }
    \label{fig:illustration}
    \vspace{-16pt}
\end{figure}

Over-sampling is an effective resampling technique through generating new synthetic samples for the minority class~\cite{krawczyk2016learning}. One of the most popular over-sampling methods in the literature is SMOTE~\cite{chawla2002smote}, which generates synthetic samples by performing linear interpolation between minority instances and their neighbors, illustrated at the top of  Figure~\ref{fig:illustration}. In contrast to the random over-sampling approach that randomly duplicates the minority instances~\cite{johnson2019survey}, SMOTE can make the decision regions larger and less specific, which could help alleviate the over-fitting issue~\cite{han2005borderline}. Despite its success, SMOTE can easily generate noisy samples since all the decisions are randomly made. For example, in Figure~\ref{fig:illustration}, one of the synthetic samples generated by SMOTE interleaves with the majorities, which could degrade the performance.

Numerous extensions have been proposed to improve SMOTE with better sampling strategies (there are at least 85 SMOTE variants as of the year of 2019~\cite{kovacs2019smote}). To name a few, ADASYN~\cite{he2008adasyn} generates more synthetic samples for the instances that are harder to learn, which is quantified by the ratio of the majority instances in the nearest neighbors. BorderlineSMOTE~\cite{han2005borderline} and SVMSMOTE~\cite{nguyen2011borderline} only over-sample the minority instances in the borderline, where the former identifies the borderline based on the nearest neighbors and the latter trains an SVM to achieve this. To avoid generating noisy samples, ANS~\cite{siriseriwan2017adaptive} proposes to adapt the number of neighbors needed for each instance based on a 1-nearest neighbor model.

However, the existing SMOTE variants heavily rely on the heuristics to perform over-sampling, which could be sub-optimal. On the one hand, the heuristic sampling strategies are often designed based on some assumptions, such as samples in the borderline are more important~\cite{han2005borderline,nguyen2011borderline}. However, the assumptions may not well hold for all the datasets and all the base classifiers, since we may need different sampling strategies in different scenarios. On the other hand, the heuristic sampling strategies cannot directly optimize the performance metric and may not deliver a desirable generalization performance. Motivated by this, we investigate the possibility of developing a learning-based over-sampling algorithm to directly optimize the performance metric. Specifically, we aim to study the following research question: \emph{Given a dataset and a base classifier, how can we optimize the over-sampling strategy such that the trained classifier can achieve the best generalization performance?}

It is non-trivial to achieve the above goal for the following challenges. \textbf{First}, it is hard to directly optimize the performance metric. The sampling is independent of the classifier so that it can only indirectly impact the performance. We need an effective mechanism to fill this gap so that the sampling strategy can be learned. \textbf{Second}, the over-sampling problem has a huge decision space since the number of generated samples can be arbitrarily large, and each synthetic sample can be anywhere in the feature space. \textbf{Third}, over-sampling is a very complex decision that requires hierarchical reasoning. At the high level, we need to decide the over-sampling ratio, i.e., how many synthetic samples should be generated. At the low level, we need to decide where the synthetic samples should be located. The low-level decision depends on the high-level decision in that the optimal locations of the samples may differ for different numbers of samples. The existing algorithms designed for similar problems, such as automated hyperparameter tuning~\cite{yu2020hyper} and neural architecture search~\cite{elsken2019neural}, often focus on a flat and much simpler search space so that they cannot be directly applied to model the potential interaction effect of the different levels of decisions in the over-sampling problem. We need a tailored algorithm to jointly optimize the hierarchical decisions to achieve the best performance.

To tackle the above challenges, we propose AutoSMOTE, an automated over-sampling algorithm that defines the search space based on SMOTE and leverages deep reinforcement learning (RL) to optimize the generalization performance, illustrated at the bottom of Figure~\ref{fig:illustration}. Motivated by the success of SMOTE and its extensions, we define a hierarchical search space based on the generation process of SMOTE to reduce the decision space. At the high level, we go through the instances one by one and decide how many samples will be generated around the current instance. At the low level, we decide which neighbors to perform linear interpolation and the interpolation weight to generate a new sample. The high-level policy is further decomposed into a cross-instance sub-policy for predicting an overall over-sampling ratio for all the instances and an instance-specific sub-policy for making personalized decisions for each instance. Then we formulate this hierarchical search space as a Markov decision process (MDP), where the three levels of the policies collaboratively make decisions. We leverage deep hierarchical RL to solve the MDP and jointly train the three levels of the policies to optimize the reward, which is obtained by the performance metric on the validation data. Extensive experiments on six real-world datasets demonstrate the superiority of AutoSMOTE. To summarize, we make the following contributions.


\begin{itemize}
    \item Formally define the problem of automated over-sampling for imbalanced classification.
    \item Define a hierarchical search space for this problem based on the generation process of SMOTE. The designed search space can cover all the possible generated samples by SMOTE and most of the SMOTE extensions.
    \item Propose AutoSMOTE for automated over-sampling. We formulate the decision process as an MDP and leverage deep hierarchical RL to directly optimize the generalization performance. We also present an implementation that runs the sampling and RL training in parallel on CPU and GPU.
    \item Conduct extensive experiments to evaluate AutoSMOTE. We show that AutoSMOTE outperforms the state-of-the-art samplers under different configurations of imbalanced ratios and base classifiers. In addition, we present comprehensive hyperparameter and ablation studies.
\end{itemize}
\begin{figure}
    \centering
    \includegraphics[width=0.4\textwidth]{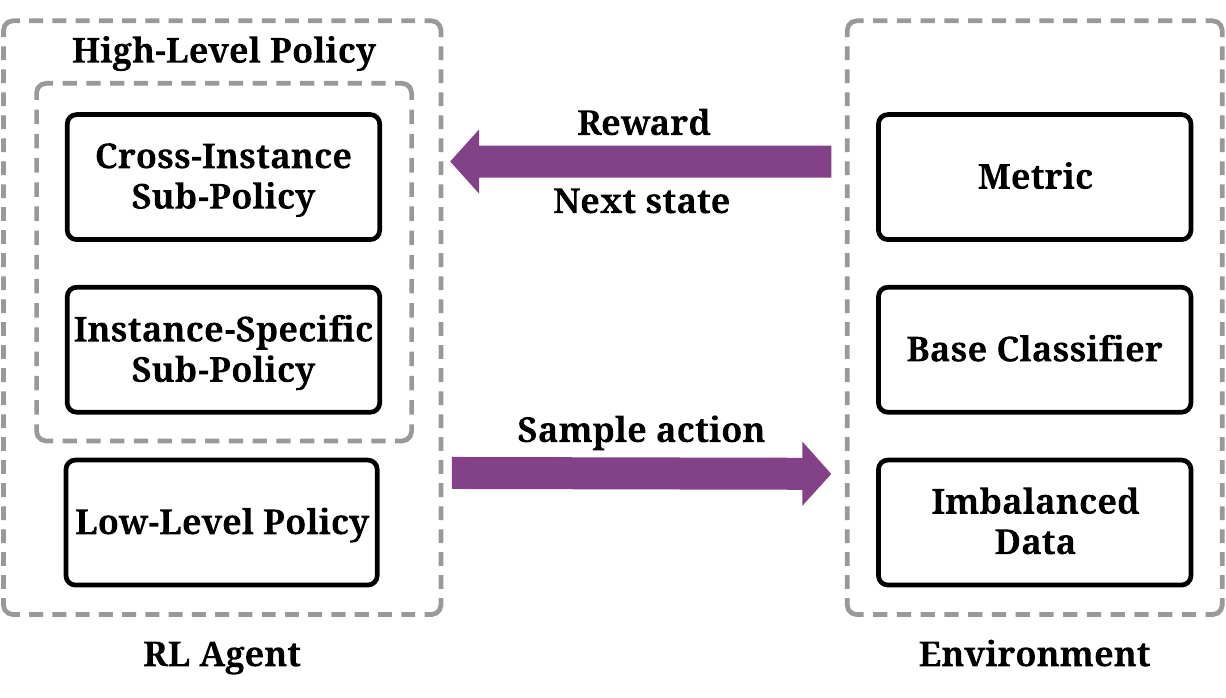}
    \vspace{-15pt}
    \caption{An overview of AutoSMOTE. The RL agent generates synthetic samples (actions) based on the current data distribution (state) with a high-level policy for deciding sampling ratios, and a low-level policy for performing actual sampling, where the high-level policy consists of two sub-policies that collaboratively make decisions. The environment takes as input the action and transits to the next state. The performance metric of the base classifier on the validation data serves as the reward to update the RL agent.}
    \vspace{-15pt}
    \label{fig:overview}
\end{figure}

\section{Problem Statement}

We focus on binary imbalanced classification problems. Let $\mathcal{X} = \{\textbf{X}^{\text{maj}}, \textbf{X}^{\text{min}}\}$ be an imbalanced dataset, where $\textbf{X}^{\text{maj}} \in \mathbb{R}^{N^{\text{maj}} \times D}$ denotes the majority instances, $\textbf{X}^{\text{min}} \in \mathbb{R}^{N^{\text{min}} \times D}$ denotes the minority instances, $N^{\text{maj}}$ is the number of majority instances, $N^{\text{min}}$ is the number of minority instances, and $D$ is the feature dimension. We define $\text{IR} = \frac{N^{\text{maj}}}{N^{\text{min}}}$ as the imbalanced ratio, where $\text{IR} > 1$. In a typical classification task, we aim to train a classifier on the training set $\mathcal{X}^{\text{train}}$, tune the performance on the validation set $\mathcal{X}^{\text{val}}$, and evaluate the trained classifier on the testing set $\mathcal{X}^{\text{test}}$. However, when $\text{IR}$ is large, the classifier may have poor performance, particularly on the minorities. Over-sampling techniques tackle this problem by augmenting $\textbf{X}^{\text{min}}$ with some synthetic samples such that the classifier can perform well on both majority and minority classes.

Based on the above notations and intuitions, we formally define the problem of automated over-sampling for imbalanced classification as follows. Given $\mathcal{X}^{\text{train}}$, $\mathcal{X}^{\text{val}}$, and a classifier $C$, we aim to generate some synthetic samples based on $\mathcal{X}^{\text{train}}$ to improve the generalization performance of $C$. Formally, the objective is to identify the best synthetic samples $\textbf{X}^{\text{syn}} \in \mathbb{R}^{N^{\text{syn}} \times D}$, where $N^{\text{syn}}$ is the number of synthetic samples, such that the performance of $C$ on $\mathcal{X}^{\text{val}}$ can be maximally improved when training $C$ on $\mathcal{X}^{\text{train}} \cup \{\textbf{X}^{\text{syn}}\}$.

\section{Methodology}

\begin{figure*}
    \centering
    \includegraphics[width=0.9\textwidth]{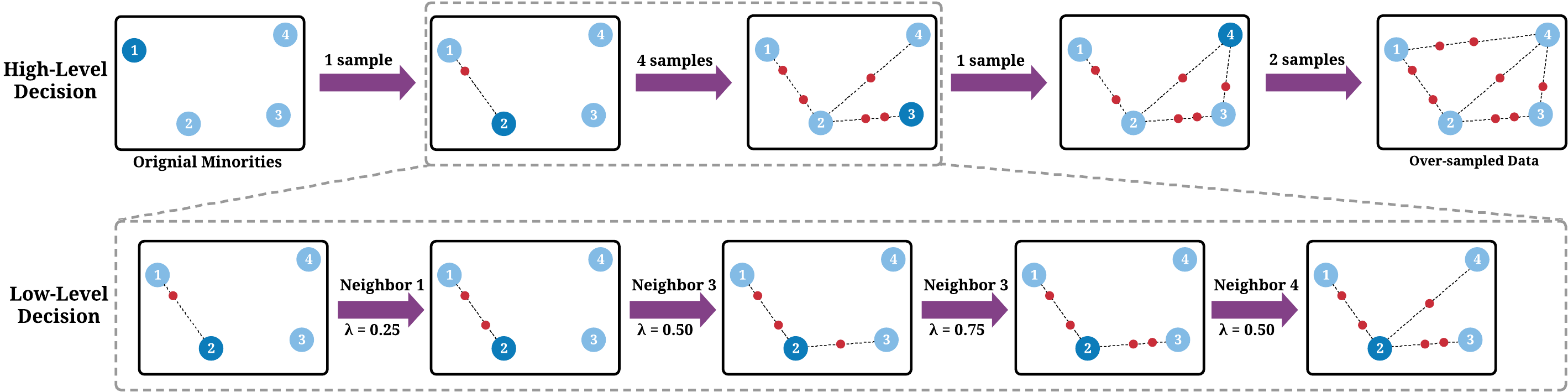}
    \vspace{-12pt}
    \caption{An illustration of the hierarchical decision process. We go through the minority instances one by one, where the darker blue instance is the current instance to be augmented. At the high level, we decide how many synthetic instances will be sampled around the current instance. At the low level, we decide which neighboring instances to perform linear interpolation and the interpolation weights $\lambda$. The low-level decision depends on the high-level decision since the length of the low-level sampling is determined by the decisions made in the high-level.}
    \vspace{-8pt}
    \label{fig:process}
\end{figure*}

Figure~\ref{fig:overview} shows an overview of AutoSMOTE. AutoSMOTE on the highest level includes an environment, which trains the base classifier on the over-sampled data and evaluates it on the validation set, and an RL agent, which learns an over-sampling strategy to optimize the performance on the validation set in a trial-and-error fashion. The RL agent consists of a high-level policy, which decides how many synthetic samples will be generated around each original training instance, and a low-level policy, which decides how the instances interpolate with the neighboring instances. The high-level policy is further decomposed into a cross-instance sub-policy for predicting an overall over-sampling ratio for all the instances and an instance-specific sub-policy for making personalized decisions of how many synthetic samples to generate for each instance. The three (sub-)policies work collaboratively to generate synthetic samples, interact with the environment, and are updated based on the reward signal. We first elaborate on the proposed hierarchical synthetic sample generation process in Section~\ref{sec:31}. Then we formulate this process as a Markov decision process~(MDP) with hierarchical policies in Section~\ref{sec:32}. Finally, in Section~\ref{sec:33}, we introduce how to optimize the MDP with hierarchical RL and present a practical implementation accelerated by multi-processing.


\vspace{-5pt}

\subsection{Hierarchical Synthetic Sample Generation}
\label{sec:31}

Given the training set $\mathcal{X}^{\text{train}}$, the goal is to generate synthetic samples $\textbf{X}^{\text{syn}} \in \mathbb{R}^{N^{\text{syn}} \times D}$ based on $\mathcal{X}^{\text{train}}$, which is a complex decision because 1) we need to determine the number of samples to generate, i.e., the value of $N^{\text{syn}}$, and 2) we need to decide where each sample is, i.e., the values of each row of $\textbf{X}^{\text{syn}}$. This leads to an extremely large search space in that there can be an arbitrary number of samples, and each sample can be anywhere in the feature space. This subsection introduces how we reduce the search space and formulate it as a hierarchical decision process based on SMOTE~\cite{chawla2002smote}.


SMOTE is one of the most popular over-sampling techniques with many extensions built upon it~\cite{kovacs2019smote,buda2018systematic}. The key idea is to generate a new sample by performing linear interpolation between a minority instance and one of its nearest neighbors. Specifically, SMOTE augments the minority instances by repeatedly executing the following steps until the desired number of synthetic samples is reached: 1) randomly pick a minority instance, 2) find the nearest minority neighbors of this instance and randomly pick a neighbor, 3) perform linear interpolation between the selected instance and the neighbor to generate a new sample, where the interpolation weight is uniformly sampled in the range of $[0,1]$; that is, the new sample is a ``mixup'' of the two original instances and lies between them. We note that all the decisions in SMOTE are randomly made. As such, many heuristics have been proposed to make the sampling process more effective, such as only over-sampling the borderline samples~\cite{han2005borderline}, and generating more samples for the instances that are harder to learn~\cite{he2008adasyn}. The generated samples of the existing extensions often still fall within the generation space defined by SMOTE.

Motivated by the effectiveness of SMOTE and its extensions, we propose to formulate the search space based on SMOTE with a hierarchical decision process. Figure~\ref{fig:process} provides an illustrative example of how we augment four minority instances with a two-level decision. At the high level, we go through the instances one by one and decide how many samples will be generated around the current instance, which leads to a $4$-step decision. For each high-level step, we will make a $g$-step low-level decision to perform linear interpolation, where $g$ is the output of the high-level step. In each low-level step, we decide which neighbors to perform linear interpolation and the interpolation weight to generate a new sample. After going through all the instances, we generate $8$ synthetic samples in total.

The proposed generation process has several desirable properties. First, it can significantly reduce the search space because we only need to decide how the current instance interpolates with its neighbors rather than blindly sampling a point in the feature space. Second, it can make personalized decisions. For example, we can generate more samples for some instances and less for the other instances. Third, it can cover all the possible generated samples by SMOTE. Moreover, since the existing SMOTE extensions often follow the generation space of SMOTE, we can essentially cover the majority of the SMOTE extensions as well.


\subsection{Formulating the Generation as MDP}
\label{sec:32}
This subsection formulates the above hierarchical decision process as MDP. A naive way to achieve this is to use a flat policy. Specifically, in each step, the agent will either make a high-level decision or a low-level decision. Taking the sampling process in Figure~\ref{fig:process} as an example, a flat policy will make the following decisions sequentially. In step 1, the agent makes a high-level decision and outputs 1. In step 2, the agent switches to low-level decision and outputs neighbor 2 and $\lambda=0.25$. In step 3, the agent comes back to the high-level decision and outputs 4. In steps 4 to 7, the agent again makes low-level decisions. Eventually, the agent will take 12 steps in total to complete the generation process.

However, this flat process will make the agent hard to train for two reasons. First, the number of steps of the MDP can be extremely large for the real-world dataset. Solving a long MDP is notoriously difficult~\cite{sutton2018reinforcement}. Second, the high-level decision and low-level decision have very different action spaces. Specifically, the high-level actions are the numbers of samples to generate, while the low-level actions are neighbors and interpolation weights. It is difficult to model the two very different action spaces under a unified policy.

To address these issues, we propose to view the MDP from a hierarchical perspective. We decompose the MDP by corresponding the high-level decision with a high-level policy $\pi_h$ and the low-level decision with a low-level policy $\pi_l$, where $\pi_h$ and $\pi_l$ use the same state features and rewards with different action spaces. We define the state, high-level/low-level actions, and reward below.
\begin{itemize}
    \item \textbf{State $s$}: a vector of features describing the data distribution. We empirically use the following features: the original features of the current instance, and the data distribution features of the current instance. We will provide more details of these features after defining the state and reward.
    \item \textbf{High-level action $g$}: an integer describing the number of samples to generate for the current instance. This essentially defines the goal of the low-level policy.
    \item \textbf{Low-level action $a$}: an action that describes which neighbor to perform interpolation and the corresponding interpolation weight. We reduce the action space by only considering the top $K$ neighbors. We then discretize the interpolation weight $\lambda$ and choose it from the set $\{0, 0.25, 0.5, 0.75, 1.0\}$, which leads to $5K$ possible actions in each step. It is possible that formulating it as a continuous action will lead to better performance, which we will study in our future work.
    \item \textbf{Reward $r$}: the reward is obtained by the performance metric on the validation data. Both $\pi_h$ and $\pi_l$ will receive a zero reward in the intermediate steps and receive a reward indicating the performance metric in the final step.
\end{itemize}

The state features consist of two parts as follows.

\begin{itemize}
    \item \textbf{Original features:} We use the original features of the current instance as the first part of the state features.
    \item \textbf{Data distribution features:} These features describe how many synthetic instances have already been generated around the current instance so that the policies can make decisions based on the previously generated synthetic samples in the MDP. Specifically, we use a 10-dimensional one-hot encoding to achieve this. Whenever an instance is used for linear interpolation (i.e., the instance is either used as the starting point or selected as a neighbor), we add one to the count of this instance. Then we use 10 bins to compress the length of the feature. For example, if the count is in the range of [0, 9], the count is put into the first bin such that the first element of the feature vector is 1 with the others being 0.
\end{itemize}


The two policies generate samples as follows. Let $N^{\text{min}}$ be the total number of minority instances. The high-level policy will generate one episode with $N^{\text{min}}$ steps. In each high-level step $t_h \in \{1, 2, ..., N^{\text{min}}\}$, $\pi_h$ takes as input the current state $s_{t_h}$ and proposes the goal for the low-level policy $g_{t_h}$, i.e., how many samples to generate. Then $\pi_l$ takes as input $g_{t_h}$ and tries to accomplish this goal by taking $g_{t_h}$ steps, where in each low-level step $t_l \in \{1, 2, ..., g_{t_h}\}$, $\pi_l$ takes as input the current state $s_{t_l}$ and outputs the sampling action $a_{t_l}$. The overall generation process will result in one high-level episode with length of $N^{\text{min}}$ and $N^{\text{min}}$ low-level episodes, whose lengths are determined by the outputs of $\pi_h$. Then, we obtain a reward with generated samples and set the final steps of all the high-level and low-level episodes to be the obtained reward, while all the intermediate steps receive zero reward.

In our preliminary experiments, we find that when $\text{IR}$ is large, we often need to generate more samples. This will significantly enlarge the action space of $\pi_h$, which makes the policy harder to train. To reduce the action space, we further decompose $\pi_h$ into a cross-instance sub-policy $\pi^{(1)}_h$ and an instance-specific sub-policy $\pi^{(2)}_h$, where $\pi^{(1)}_h$ and $\pi^{(2)}_h$ use the same state features and rewarding scheme but differs in the action space. Specifically, $\pi^{(1)}_h$ only takes one step in the generation process and outputs a cross-instance scaling factor $g^{(1)} \in \{0,1,...,G^{(1)}\}$, where $G^{(1)}$ is a hyperparameter. $\pi^{(2)}_h$ performs $N^{\text{min}}$ steps and outputs an instance-specific scaling factor $g^{(2)} \in \{0,1,...,G^{(2)}\}$ for each instance, where $G^{(2)}$ is a hyperparameter. Then the high-level action is obtained by $g = g^{(1)} \times g^{(2)}$. Algorithm 1 summarizes how $\pi^{(1)}_h$, $\pi^{(2)}_h$ and $\pi_l$ collaboratively interact with the environment to generate samples.

The above design of three-level hierarchical policies enjoys several advantages. First, it can significantly reduce the length of the episodes. Specifically, the episode length of the $\pi^{(1)}_h$ is only 1, the episode length of $\pi^{(2)}_h$ is $N^{\text{min}}$, and the length of each low-level episode is determined by $g$, all of which are much smaller than that of the flat counterpart. Second, by further decomposing the high-level policy, we can significantly reduce the action space from $G^{(1)} \times G^{(2)}$ to $G^{(1)}$ and $G^{(2)}$ for the two sub-policies. Third, each level of the hierarchy plays a different role in the decision, where $\pi^{(1)}_h$ makes the dataset-level decisions of the desired over-sampling ratio, $\pi^{(2)}_h$ makes personalized decisions to allow generating more samples for some instances, and $\pi_l$ performs actual sampling based on the specified goals. As such, we can naturally model them with three separate policies to learn these three very different skills.

\begin{algorithm}[t]
\caption{Generation process of AutoSMOTE}
\label{alg:1}
\setlength{\intextsep}{0pt} 
\begin{algorithmic}[1]
\STATE \textbf{Input:} cross-instance sub-policy $\pi^{(1)}_h$, instance-specific sub-policy $\pi^{(2)}_h$, low-level policy $\pi_l$, minority instances $\textbf{X}^{\text{min}} \in \mathbb{R}^{N^{\text{min}} \times D}$, max cross-instance scaling factor $G_1$, max instance-specific scaling factor $G_2$, max number of neighbors $K$
\STATE Sample $g^{(1)}$ in the set of $\{0,1,...,G^{(1)}\}$ with $\pi^{(1)}_h$
\FOR{instance ID = 1, 2, ..., $N^{\text{min}}$}
    \STATE Sample $g^{(2)}$ in the set of $\{0,1,...,G^{(2)}\}$ with $\pi^{(2)}_h$
    \STATE $g \leftarrow g^{(1)} \times g^{(2)}$
    \FOR{iteration = 1, 2, ..., $g$}
        \STATE Sample a top-$K$ neighbor of the current instance and an interpolation weight with $\pi_l$, and generate a new sample
    \ENDFOR
\ENDFOR
\end{algorithmic}
\end{algorithm}

\subsection{Optimizing the Generation Process with Deep Hierarchical RL}
\label{sec:33}
This subsection introduces how we optimize the three-level hierarchical policies with deep hierarchical RL. We first describe the training objectives for the three policies. Then we present the overall training procedure accelerated by multi-processing. 

To enable the training of the policies with gradient descent, we parameterize the three policies. Following the idea of actor-critic~\cite{sutton2018reinforcement}, we associate each of the policies with a policy-value network. For $\pi^{(1)}_h$, we first use an MLP to process the state and produce a state representation. Then the state representation is processed by a policy head and a value head, where the policy head is a fully-connected layer followed by a Softmax layer to produce action probabilities $\pi^{(1)}_h(g^{(1)}|s)$, and the value head $V^{(1)}_h(s)$ is a fully-connected layer with output dimension of 1. We use the same procedure to process $\pi^{(2)}_h$ to obtain $\pi^{(2)}_h(g^{(2)}|s)$ and $V^{(2)}_h(s)$. For $\pi_l$, the value head $V_l(s)$ is obtained in the same way as above. For the policy head of $\pi_l$, we extract action features, which include data distribution and interpolation weight. The data distribution features are extracted in the same way as the state features, and the interpolation weight features are obtained by one-hot encoding. The action features are then concatenated with the state representation, followed by an MLP to produce confidence scores for state-action pairs. Finally, a Softmax layer is applied to all the state-action pairs to produce action probabilities $\pi_l(a|s)$.

\begin{algorithm}[t]
\caption{Training of AutoSMOTE}
\label{alg:2}
\setlength{\intextsep}{0pt} 
\begin{algorithmic}[1]
\STATE \textbf{Input:} $\pi^{(1)}_h$, $\pi^{(2)}_h$, $\pi_l$, $\textbf{X}^{\text{min}}$, $G_1$, $G_2$, $K$, total number of iterations $I$, three buffer sizes $B^{(1)}_h$, $B^{(2)}_h$, and $B_l$ 
\STATE Initialize three queue buffers $\mathcal{B}^{(1)}_h$, $\mathcal{B}^{(2)}_h$, $\mathcal{B}_l$
\FOR{iteration = 1, 2, ..., $I$}
    \STATE Generate samples following Algorithm~\ref{alg:1} and store the generated episodes to $\mathcal{B}^{(1)}_h$, $\mathcal{B}^{(2)}_h$ and $\mathcal{B}_l$
    \STATE Train on the augmented training data, get reward on validation data, and set the final steps of all the episodes to be the obtained reward with all the intermediate rewards as 0
    \IF{$\text{size}(\mathcal{B}^{(1)}_h) \ge B^{(1)}_h$}
        \STATE Pop out $B^{(1)}_h$ steps of data and update $\pi^{(1)}_h$ with Eq.~\ref{eq:2}
    \ENDIF
    \IF{$\text{size}(\mathcal{B}^{(2)}_h) \ge B^{(2)}_h$}
        \STATE Pop out $B^{(2)}_h$ steps of data and update $\pi^{(2)}_h$ with Eq.~\ref{eq:2}
    \ENDIF
    \IF{$\text{size}(\mathcal{B}_l) \ge B_l$}
        \STATE Pop out $B_l$ steps of data and update $\pi_l$ with Eq.~\ref{eq:2}
    \ENDIF
\ENDFOR
\end{algorithmic}
\end{algorithm}

We adopt the feudal hierarchy approaches~\cite{pateria2021hierarchical} to train the policies, where each level of the policies observes the environment in different granularity, and the three policies are updated simultaneously. To train the policies, we adopt IMPALA~\cite{espeholt2018impala}, a modern distributed deep RL algorithm. Here, we mainly introduce the high-level procedure since RL itself is not our focus; for readers who are interested in how IMPALA works, please refer to~\cite{espeholt2018impala}. Let $s_t$, $a_t$, and $r_t$ be the state, action, and reward at step $t$, respectively. For brevity, here we abuse the notation of $a_t$, which will actually be $g^{(1)}_t$ and $g^{(2)}_t$ in the context of $\pi^{(1)}_h$ and $\pi^{(2)}_h$, respectively. We consider an n-step trajectory $(s_t, a_t, r_t)_{t=t'}^{t=t'+n}$. IMPALA uses a V-trace target for $s_{t'}$ for tackling the delayed model update:
\begin{equation}
    v_{t'} = V(s_{t'}) + \sum_{t=t'}^{t'+n-1} \gamma^{t-t'}(\prod_{i=t'}^{t-1} c_i) \delta_t V,
    \label{eq:1}
\end{equation}
where $V(s_{t'})$ is the value head for $s_{t'}$, $\delta_t V = \rho_t (r_t + \gamma V(s_{t+1}) - V(x_t))$ is the temporal difference, and $c_i$ and $\rho_t$ are truncated importance sampling weights. The loss at step $t$ is defined as
\begin{equation}
    L_t = \rho_t \log \pi(a_t|s_t) (r_t + \gamma v_{t+1} - V(s_t)) + \frac{1}{2} (v_t - V(s_t))^2,
    \label{eq:2}
\end{equation}
where $\pi(a_t|s_t)$ is the policy head, and $V(s_t)$is the value head. Batch training will be further used to update the model for multiple steps at a time. The V-trace correction is helpful because training and evaluating a classifier may result in substantial delays. The three policies are updated simultaneously based on Eq.~\ref{eq:2}.

Algorithm~\ref{alg:2} summarizes the training procedure. We first initialize three buffers to temporally store the generated data from the environment in line 2, i.e., tuples of $\langle \text{state}, \text{action}, \text{reward} \rangle$. In each iteration, we generate new samples (line 4) and obtain rewards on the validation data (line 5). The three policies are updated periodically with the RL objectives (lines 6 to 14). The generated samples with the highest validation performance will be used to re-train the classifier, which will be further evaluated on a hold-out testing set. 

\textbf{Multi-Processing Implementation.} To maximally exploit the computational resources, we run the sampling (lines 4 and 5) with multiple processes in the CPU and do the learning (lines 6 to 14) in the main process with GPU. We find that the sampling part is CPU-intensive because the base classifier is trained with CPU. In contrast, the learning part can be accelerated with GPU since updating the model can be accelerated by parallelism. Modern GPU servers often have multiple CPU cores. This motivates us to accelerate the training with multi-processing. Specifically, we run multiple actors, where each actor runs a separate clasffier to sample data. Then we run a single learner on GPU to update the model weights with the data collected from the actors. The actors and the learner communicate with shared memory. 



\begin{table}[t]
    \centering
    \footnotesize
    \caption{Dataset statistics with imbalanced ratios of 20/50/100.}
    \vspace{-10pt}
    \label{tab:stats}
    \setlength{\tabcolsep}{4pt}
    \begin{tabular}{l|cccc}
    \toprule
     
  & \# Majorities & \# Minorities & \# Features & Domain \\
 
    \midrule
    \midrule
     Phoneme & 3818 & 190/76/38 & 5 & Audio \\
     PhishingWebsites & 6157 & 307/123/61 & 68 & Security \\
     EEGEyeState & 8257 & 412/165/82 & 14 & EEG \\
     Mozilla4 & 10437 & 521/208/104 & 5 & Product defect \\
     MagicTelescope & 12332 & 616/246/123 & 10 & Telescope \\
     Electricity & 26075 & 1303/521/260 & 14 & Electricity \\
     \bottomrule
    \end{tabular}
    \vspace{-15pt}
\end{table}

\section{Experiments}

The experiments aim to answer the following research questions. \textbf{RQ1:} How does AutoSMOTE compare with the state-of-the-art techniques for imbalanced classification~(Sections~\ref{sec:42})? \textbf{RQ2:} Can AutoSMOTE outperform the numerous variants of SMOTE~(Sections~\ref{sec:43})? \textbf{RQ3:} How does each component contribute to the performance of AutoSMOTE and how does AutoSMOTE compare with simple random search~(Sections~\ref{sec:44})? \textbf{RQ4:} How do the hyperparameters impact the performance of AutoSMOTE~(Sections~\ref{sec:45})? \textbf{RQ5:} How does the learned balancing strategy of AutoSMOTE compare with the existing over-sampling strategies~(Sections~\ref{sec:46})?


\begin{table*}[t]
    \centering
    \footnotesize
    \caption{Average ranks (the lower the better) of AutoSMOTE and baselines in terms of Macro-F1/MCC. We use $\blacktriangle$ to denote the cases where AutoSMOTE is significantly better than baselines w.r.t. the Wilcoxon signed rank test (p < 0.05).}
    \vspace{-10pt}
    \label{tab:performance}
    \setlength{\tabcolsep}{2pt}
    \begin{tabular}{l|l|cccccc|c}
    \toprule
     
    \multirow{2}{*}{Category} & \multirow{2}{*}{Method} & \multicolumn{6}{c|}{Dataset} & \multirow{2}{*}{Overall} \\
    \cline{3-8}
    ~ & ~ & Phoneme & PhishingWebsites & EEGEyeState & Mozilla4 & MagicTelescope  & Electricity & ~\\
    \midrule
    \midrule

     No-resampling & - & 16.50$\blacktriangle$/16.75$\blacktriangle$ & 9.75$\:\:\:\:\;$/9.42$\:\:\:\:\;$ & 16.08$\blacktriangle$/17.08$\blacktriangle$ & 12.75$\blacktriangle$/12.83$\blacktriangle$ & 15.92$\blacktriangle$/13.00$\blacktriangle$ & 18.17$\blacktriangle$/15.67$\blacktriangle$ & 14.86$\blacktriangle$/14.12$\blacktriangle$\\
     \midrule
     \multirow{11}{*}{Under-sampling} & ClusterCentroids & 13.25$\blacktriangle$/14.58$\blacktriangle$ & 19.50$\blacktriangle$/19.58$\blacktriangle$ & 13.67$\blacktriangle$/14.33$\blacktriangle$ & 15.67$\blacktriangle$/15.42$\blacktriangle$ & 16.42$\blacktriangle$/19.58$\blacktriangle$ & 15.25$\blacktriangle$/18.25$\blacktriangle$ & 15.62$\blacktriangle$/14.58$\blacktriangle$ \\
     ~ & CondensedNearestNeighbour & 16.62$\blacktriangle$/17.46$\blacktriangle$ & 16.92$\blacktriangle$/16.96$\blacktriangle$ & 17.75$\blacktriangle$/19.17$\blacktriangle$ & 19.88$\blacktriangle$/20.46$\blacktriangle$ & 14.33$\blacktriangle$/14.92$\blacktriangle$ & 13.58$\blacktriangle$/14.58$\blacktriangle$ & 16.51$\blacktriangle$/17.19$\blacktriangle$ \\
     ~ & EditedNearestNeighbours & 14.17$\blacktriangle$/15.42$\blacktriangle$ & 11.83$\:\:\;$/12.42$\:\:\;$ & 15.71$\blacktriangle$/16.88$\blacktriangle$ & 11.96$\blacktriangle$/12.46$\blacktriangle$ & 13.04$\blacktriangle$/10.71$\:\:\;$ & 15.25$\blacktriangle$/15.50$\blacktriangle$ & 13.66$\blacktriangle$/13.90$\blacktriangle$ \\
     ~ & RepeatedEditedNearestNeighbours & 14.71$\blacktriangle$/17.04$\blacktriangle$ & 15.29$\blacktriangle$/15.96$\blacktriangle$ & 15.33$\blacktriangle$/17.17$\blacktriangle$ & 13.88$\blacktriangle$/14.12$\blacktriangle$ & 10.38$\blacktriangle$/8.88$\:\:\:\:\;$ & 14.62$\blacktriangle$/14.88$\blacktriangle$ & 14.03$\blacktriangle$/14.67$\blacktriangle$ \\
     ~ & AllKNN & 14.04$\blacktriangle$/15.46$\blacktriangle$ & 13.46$\blacktriangle$/13.46$\:\:\;$ & 15.50$\blacktriangle$/16.75$\blacktriangle$ & 13.88$\blacktriangle$/14.29$\blacktriangle$ & 10.71$\blacktriangle$/8.38$\:\:\:\:\;$ & 16.08$\blacktriangle$/17.17$\blacktriangle$ & 13.94$\blacktriangle$/14.25$\blacktriangle$ \\
     ~ & InstanceHardnessThreshold & 14.21$\blacktriangle$/13.38$\blacktriangle$ & 20.67$\blacktriangle$/20.67$\blacktriangle$ & 14.29$\blacktriangle$/14.46$\blacktriangle$ & 17.79$\blacktriangle$/18.04$\blacktriangle$ & 10.25$\:\:\;$/10.25$\:\:\;$ & 12.58$\blacktriangle$/12.83$\blacktriangle$ & 14.97$\blacktriangle$/14.94$\blacktriangle$ \\
     ~ & NearMiss & 24.42$\blacktriangle$/24.75$\blacktriangle$ & 24.17$\blacktriangle$/24.42$\blacktriangle$ & 22.58$\blacktriangle$/20.92$\blacktriangle$ & 23.25$\blacktriangle$/23.33$\blacktriangle$ & 25.00$\blacktriangle$/25.00$\blacktriangle$ & 19.58$\blacktriangle$/21.08$\blacktriangle$ & 23.17$\blacktriangle$/23.25$\blacktriangle$ \\
     ~ & NeighbourhoodCleaningRule & 16.33$\blacktriangle$/17.83$\blacktriangle$ & 13.08$\:\:\;$/13.25$\blacktriangle$ & 15.12$\blacktriangle$/15.88$\:\:\;$ & 12.83$\blacktriangle$/13.25$\blacktriangle$ & 10.79$\:\:\;$/9.12$\:\:\blacktriangle$ & 12.54$\blacktriangle$/11.71$\blacktriangle$ & 13.45$\blacktriangle$/13.51$\blacktriangle$ \\
     ~ & OneSidedSelection & 17.21$\blacktriangle$/18.21$\blacktriangle$ & 11.08$\:\:\;$/10.67$\blacktriangle$ & 15.92$\blacktriangle$/16.17$\blacktriangle$ & 13.83$\blacktriangle$/14.17$\blacktriangle$ & 15.62$\blacktriangle$/13.21$\blacktriangle$ & 17.12$\blacktriangle$/15.71$\blacktriangle$ & 15.13$\blacktriangle$/14.69$\blacktriangle$\\
     ~ & RandomUnderSampler & 12.50$\blacktriangle$/10.00$\blacktriangle$ & 17.42$\blacktriangle$/17.58$\blacktriangle$ & 11.00$\blacktriangle$/9.25$\:\:\blacktriangle$ & 12.83$\blacktriangle$/12.92$\blacktriangle$ & 10.17$\:\:\;$/11.67$\:\:\;$ & 10.83$\blacktriangle$/12.25$\blacktriangle$ & 12.46$\blacktriangle$/12.28$\blacktriangle$ \\
     ~ & TomekLinks & 16.88$\blacktriangle$/17.46$\blacktriangle$ & 10.04$\:\:\;$/9.38$\:\:\:\:\;$ & 14.12$\blacktriangle$/14.04$\blacktriangle$ & 14.12$\blacktriangle$/14.29$\blacktriangle$ & 15.62$\blacktriangle$/12.96$\blacktriangle$ & 17.38$\blacktriangle$/16.21$\blacktriangle$ & 14.69$\blacktriangle$/14.06$\blacktriangle$ \\
     \midrule
     \multirow{7}{*}{Over-sampling} & RandomOverSampler & 6.75$\:\:\:\:\;$/8.17$\:\:\:\:\;$ & 12.33$\blacktriangle$/12.75$\blacktriangle$ & 5.00$\:\:\:\:\;$/5.58$\:\:\:\:\;$ & 8.00$\:\:\blacktriangle$/8.42$\:\:\blacktriangle$ & 9.92$\:\:\blacktriangle$/13.33$\:\:\;$ & 8.58$\:\:\blacktriangle$/10.83$\blacktriangle$ & 8.43$\:\:\blacktriangle$/9.85$\:\:\blacktriangle$ \\
     ~ & SMOTE & 7.25$\:\:\:\:\;$/8.67$\:\:\blacktriangle$ & 10.42$\blacktriangle$/10.67$\blacktriangle$ & 7.00$\:\:\blacktriangle$/6.67$\:\:\:\:\;$ & 12.00$\blacktriangle$/12.00$\blacktriangle$ & 11.42$\blacktriangle$/14.17$\blacktriangle$ & 7.17$\:\:\blacktriangle$/7.42$\:\:\blacktriangle$ & 9.21$\:\:\blacktriangle$/9.93$\:\:\blacktriangle$ \\
     ~ & SMOTEN & 16.83$\blacktriangle$/18.25$\blacktriangle$ & 10.71$\:\:\;$/10.54$\:\:\;$ & 12.42$\blacktriangle$/15.58$\blacktriangle$ & 9.58$\:\:\blacktriangle$/10.08$\:\:\;$ & 18.17$\blacktriangle$/17.33$\blacktriangle$ & 17.83$\blacktriangle$/18.67$\blacktriangle$ & 14.26$\blacktriangle$/15.08$\blacktriangle$ \\
     ~ & ADASYN & 7.33$\:\:\:\:\;$/8.00$\:\:\:\:\;$ & 9.75$\:\:\:\:\;$/9.58$\:\:\:\:\;$ & 7.50$\:\:\blacktriangle$/8.17$\:\:\blacktriangle$ & 12.58$\blacktriangle$/12.25$\blacktriangle$ & 10.17$\:\:\;$/12.08$\blacktriangle$ & 8.00$\:\:\blacktriangle$/8.50$\:\:\blacktriangle$ & 9.22$\:\:\blacktriangle$/9.76$\:\:\blacktriangle$ \\
     ~ & BorderlineSMOTE & 6.92$\:\:\:\:\;$/8.67$\:\:\:\:\;$ & 9.42$\:\:\blacktriangle$/9.25$\:\:\:\:\;$ & 9.67$\:\:\blacktriangle$/10.92$\blacktriangle$ & 9.17$\:\:\blacktriangle$/9.33$\:\:\blacktriangle$ & 7.50$\:\:\:\:\;$/9.75$\:\:\:\:\;$ & 4.67$\:\:\:\:\;$/5.08$\:\:\:\:\;$ & 7.89$\:\:\blacktriangle$/8.83$\:\:\blacktriangle$ \\
     ~ & KMeansSMOTE & 15.92$\blacktriangle$/16.67$\blacktriangle$ & 10.00$\:\:\;$/9.83$\:\:\:\:\;$ & 16.08$\blacktriangle$/16.92$\blacktriangle$ & 12.83$\blacktriangle$/12.79$\blacktriangle$ & 14.92$\blacktriangle$/12.17$\:\:\;$ & 17.92$\blacktriangle$/15.42$\blacktriangle$ & 14.61$\blacktriangle$/13.97$\blacktriangle$ \\
     ~ & SVMSMOTE & 6.25$\:\:\:\:\;$/9.08$\:\:\blacktriangle$ & 10.17$\blacktriangle$/10.00$\:\:\;$ & 7.25$\:\:\:\:\;$/7.75$\:\:\:\:\;$ & 8.25$\:\:\blacktriangle$/9.33$\:\:\blacktriangle$ & 6.67$\:\:\:\:\;$/8.58$\:\:\:\:\;$ & 4.50$\:\:\:\:\;$/4.83$\:\:\:\:\;$ & 7.18$\:\:\blacktriangle$/8.26$\:\:\blacktriangle$ \\
      \midrule
      Combined over- and & SMOTEENN & 6.25$\:\:\:\:\;$/6.50$\:\:\:\:\;$ & 14.67$\:\:\;$/14.50$\:\:\;$ & 7.17$\:\:\:\:\;$/6.92$\:\:\:\:\;$ & 10.75$\blacktriangle$/10.08$\:\:\;$ & 8.00$\:\:\:\:\;$/7.42$\:\:\:\:\;$ & 9.75$\:\:\blacktriangle$/9.67$\:\:\blacktriangle$ & 9.43$\:\:\blacktriangle$/9.18$\:\:\blacktriangle$ \\
      under-sampling & SMOTETomek & 8.67$\:\:\:\:\;$/9.25$\:\:\blacktriangle$ & 9.58$\:\:\:\:\;$/9.75$\:\:\:\:\;$ & 6.67$\:\:\blacktriangle$/6.42$\:\:\:\:\;$ & 11.42$\blacktriangle$/11.08$\blacktriangle$ & 8.58$\:\:\:\:\;$/10.25$\:\:\;$ & 7.75$\:\:\blacktriangle$/7.92$\:\:\blacktriangle$ & 8.78$\:\:\blacktriangle$/9.11$\:\:\blacktriangle$ \\
      \midrule
     \multirow{2}{*}{Generative models} & CTGAN & 12.08$\:\:\;$/9.33$\:\:\blacktriangle$ & 11.75$\blacktriangle$/11.42$\blacktriangle$ & 11.50$\blacktriangle$/12.58$\:\:\;$ & 10.42$\blacktriangle$/9.25$\:\:\blacktriangle$ & 15.17$\:\:\;$/15.42$\blacktriangle$ & 12.75$\:\:\;$/11.83$\blacktriangle$ & 12.28$\blacktriangle$/11.64$\blacktriangle$ \\
     ~ & TVAE & 14.25$\blacktriangle$/12.17$\blacktriangle$ & 9.42$\:\:\:\:\;$/9.17$\:\:\:\:\;$ & 23.50$\blacktriangle$/18.50$\blacktriangle$ & 16.17$\blacktriangle$/16.58$\blacktriangle$ & 20.92$\blacktriangle$/20.92$\blacktriangle$ & 19.58$\blacktriangle$/17.25$\blacktriangle$ & 17.31$\blacktriangle$/15.76$\blacktriangle$ \\
     \midrule
     Meta-learning & MESA & 19.92$\blacktriangle$/7.33$\:\:\:\:\;$ & 17.08$\blacktriangle$/16.50$\blacktriangle$ & 20.17$\blacktriangle$/12.17$\blacktriangle$ & 17.50$\blacktriangle$/13.25$\blacktriangle$ & 19.58$\blacktriangle$/18.92$\blacktriangle$ & 20.83$\blacktriangle$/18.50$\blacktriangle$ & 19.18$\blacktriangle$/14.44$\blacktriangle$ \\
     \midrule
     Auto-sampling & AutoSMOTE & \textbf{5.75}$\:\:\:\;$/\textbf{4.58}$\:\:\:\;$ & \textbf{6.50}$\:\:\:\;$/\textbf{7.67}$\:\:\:\;$ & \textbf{4.00}$\:\:\:\;$/\textbf{4.75}$\:\:\:\;$ & \textbf{3.67}$\:\:\:\;$/\textbf{4.96}$\:\:\:\;$ & \textbf{5.75}$\:\:\:\;$/\textbf{7.00}$\:\:\:\;$ & \textbf{2.67}$\:\:\:\;$/\textbf{3.25}$\:\:\:\;$ & \textbf{4.72}$\:\:\:\;$/\textbf{5.37}$\:\:\:\;$ \\
   
     \bottomrule
    \end{tabular}
    \vspace{-10pt}
\end{table*}

\subsection{Experimental Settings}
\label{sec:41}

\textbf{Datasets.} The experiments are conducted on six binary classification datasets: \textbf{Phoneme}, \textbf{PhishingWebsites}, \textbf{EEGEyeState}, \textbf{Mozilla4}, \textbf{MagicTelescope}, and \textbf{Electricity}. All the datasets are publicly available at OpenML\footnote{\url{https://www.openml.org/}}\cite{vanschoren2014openml}. We perform the following pre-processing steps. First, we identify the numerical features and the categorical features from the datasets. Second, we scale the numerical features with \texttt{StardardScaler} in sklearn, which subtracts each value with the mean value and scales the resulting value with the standard deviation. Third, we impute the missing values with 0 for all the features. Fourth, for the categorical features, we use a one-hot encoding. Since these datasets are relatively balanced, following the previous work~\cite{wang2020global,buda2018systematic,zhao2021graphsmote}, we artificially create the imbalanced datasets by randomly under-sampling the minority class with different imbalanced ratios. Then we randomly split 60\%/20\%/20\% of the data as the training/validation/testing sets, respectively. Table~\ref{tab:stats} summarizes the statistics of the imbalanced datasets.



\textbf{Baselines.} First, we include all the \textbf{under-sampling}, \textbf{over-sampling}, and \textbf{combined over- and under-sampling} methods provided in Imbalanced-learn\footnote{\url{https://imbalanced-learn.org/}}. Second, we consider two state-of-the-art \textbf{generative models} designed for generating realistic samples based on variational autoencoder (TVAE) and generative adversarial networks (CTGAN)~\cite{xu2019modeling}. We leverage TVAE and CTGAN to augment the minority class. Third, we involve MESA~\cite{liu2020mesa}, a state-of-the-art \textbf{meta-learning} algorithm, which similarly maximizes the validation performance with ensemble learning of under-sampling strategies. Finally, we compare AutoSMOTE with 85 SMOTE variants provided in Smote-variants package\footnote{\url{https://github.com/analyticalmindsltd/smote_variants}}~\cite{kovacs2019smote} in Section~\ref{sec:43}.

\textbf{Evaluation Metric.} Following the previous work of imbalanced classification~\cite{rout2018handling,johnson2019survey,boughorbel2017optimal}, we use Macro-F1 and Matthews Correlation Coefficient (MCC) to evaluate the performance. Macro-F1 calculates the F-measure separately for each class and averages them. MCC takes into account true and false positives and negatives. Both of them can well reflect the performance of the minority class.

\textbf{Base Classifiers and Imbalanced Ratios.} The performance of a resampling algorithm is highly sensitive to the adopted base classifier and the imbalanced ratio of the dataset~\cite{kovacs2019empirical,liu2020mesa}. An algorithm that performs well under one configuration may not necessarily perform well under another configuration. For a comprehensive evaluation, we rank the samplers under 12 configurations with four representative classifiers, including SVM, KNN, DecisionTree and AdaBoost, and with three imbalanced ratios of 20, 50, and 100. We report the average ranks of Macro-F1 or MCC of the samplers across the 12 configurations for each dataset and the overall ranks across all the datasets and configurations. In this way, a higher ranked algorithm tends to perform well under different base classifiers and imbalanced ratios. We run all the experiments five times and report the average results. We further employ Wilcoxon signed rank test~\cite{wiki:Wilcoxon_signed-rank_test} to rigorously compare the samplers.

\textbf{Guidance of Validation Set.} For a fair comparison, \emph{all the samplers (including all the baselines) leverage the validation set to search the sampling strategy or tune hyperparameters based on the performance on the validation set}: \textbf{1)} for AutoSMOTE, we store the over-sampled data with the best validation performance in search. Then we use the classifier trained on the stored data for evaluation. \textbf{2)} for MESA, we similarly use the performance on the validation data to train the ensemble strategy. \textbf{3)} for the other baselines, we grid-search the desired ratio of the number of minority samples over the number of majority samples after resampling (which is often a dominant hyperparameter and is suggested to be tuned by~\cite{kovacs2019empirical}) in the set of $\{0.1, 0.2, 0.3, 0.4, 0.5, 0.6, 0.7, 0.8, 0.9, 1.0\}$ on the validation set. For all the samplers, we use the best configuration discovered on the validation set and report the results on the testing set, which is unseen in training or tuning the samplers.

\textbf{Implementation Details.} For AutoSMOTE, we set the max instance-specific scaling factor $G_2=10$, the max instance-specific scaling factor $G_1=4 \times \text{IR} / G_1$ such that $G_1 \times G_2$ is 4 times of the imbalanced ratio, the max number of neighbors $K=30$, the total number of iterations $I=1000$, the buffer sizes $B^{(1)}_h=2$, $B^{(1)}_h=300$, $B_l=300$. We adopt Adam optimizer with a learning rate of 0.005. All the three policies use 128-128 MLP.  We run 40 actors in parallel to train the base classifiers. For the baselines, we use authors' implementation\footnote{\url{https://github.com/ZhiningLiu1998/mesa}} with the default hyperparameters except that we change the metric to Macro-F1 and MCC. For TVAE and CTGAN, we use the authors' implementation\footnote{\url{https://github.com/sdv-dev/CTGAN}} with the default hyperparameters. We use NVIDIA GeForce RTX 2080 Ti GPUs.


\begin{figure}[t]
  \centering
  \begin{subfigure}[b]{0.2\textwidth}
    \centering
    \includegraphics[width=0.99\textwidth]{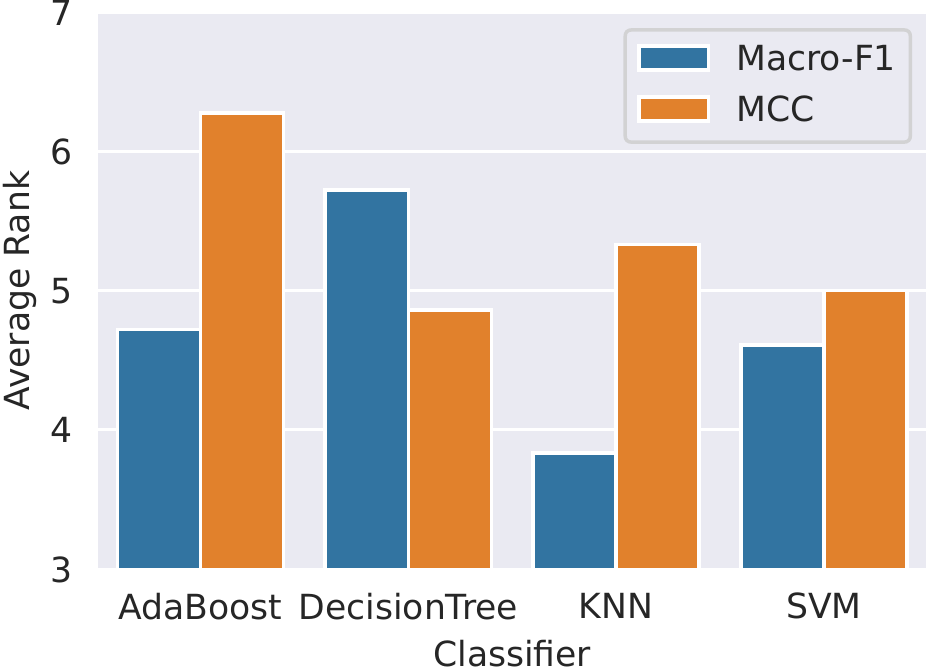}
  \end{subfigure}%
  \begin{subfigure}[b]{0.2\textwidth}
    \centering
    \includegraphics[width=0.99\textwidth]{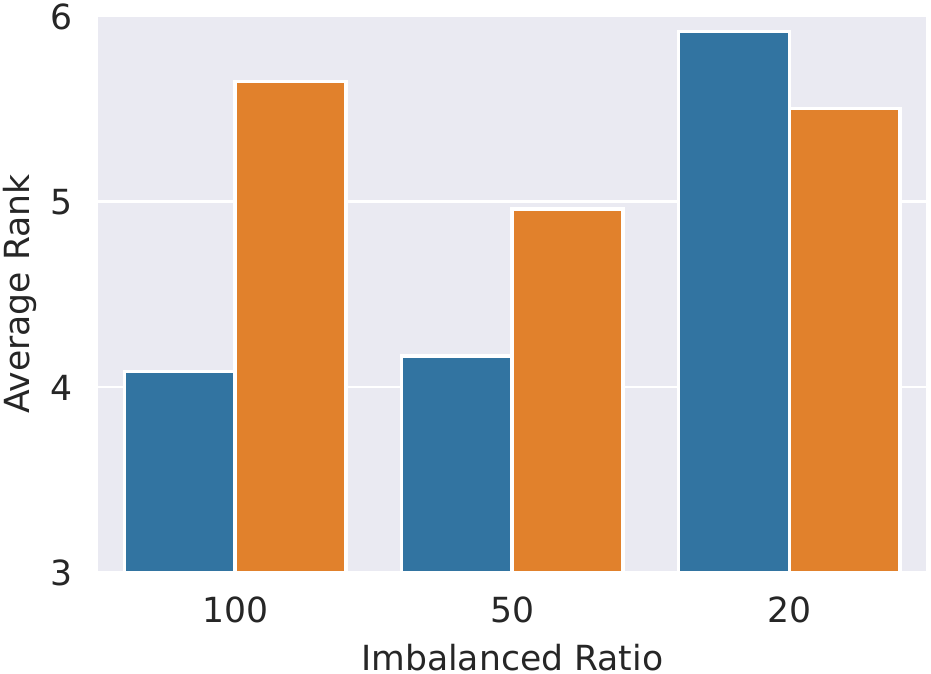}
  \end{subfigure}%
  \vspace{-10pt}
  \caption{Performance of AutoSMOTE with different base classifiers (left) and imbalanced ratios (right).}
  \vspace{-10pt}
  \label{fig:compare}
\end{figure}

\subsection{Comparison with the Baselines}
\label{sec:42}
To answer \textbf{RQ1}, we compare AutoSMOTE against the state-of-the-art resampling methods for imbalanced classification. Table~\ref{tab:performance} reports the overall ranks of Macro-F1 and MCC of the samplers on each of the datasets across all the configurations of base classifier and imbalanced ratio. We further report the overall ranks of AutoSMOTE under different classifiers and imbalanced ratios in Figure~\ref{fig:compare} to show insights into how different configurations impact the performance. We make the following observations.

First, AutoSMOTE significantly and consistently outperforms all the baselines across all the datasets. Specifically, AutoSMOTE is ranked the top among the 25 samplers for all the datasets, and the overall ranks across the datasets are significantly better than all the baselines w.r.t. Wilcoxon signed rank test. This demonstrates the superiority of the RL-based sampling strategy. An interesting observation is that while MESA also performs searching on the validation set, it cannot deliver competitive performance as AutoSMOTE. A possible explanation is that MESA searches the ensemble strategies of under-sampling, which could lose information. This phenomenon can also be verified from the observation that the over-sampling methods outperform the under-sampling methods in general for all the datasets. AutoSMOTE performs better than the two generative models, which is because AutoSMOTE can optimize the generalization performance with RL while the generative models can only model the data distribution. This also suggests that the SMOTE search space is effective since we can identify very strong synthetic strategies within the search space.

Second, AutoSMOTE delivers consistent performance across different classifiers and imbalanced ratios. The average rank of AutoSMOTE is better than 6.3 across all the classifiers and better than 6 across all the imbalanced ratios. The results suggest that AutoSMOTE can well accommodate different configurations. A possible reason is that AutoSMOTE can identify personalized synthetic strategies for different configurations with RL.

\begin{table}[t]
    \centering
    \footnotesize
    \caption{AutoSMOTE versus 85 SMOTE variants. We only list the average ranks of the top 5 algorithms due to space limitation. $\blacktriangle$ suggests AutoSMOTE is significantly better.}
    \vspace{-10pt}
    \label{tab:variants}
    \setlength{\tabcolsep}{10pt}
    \begin{tabular}{l|c||l|c}
    \toprule
     
  Sampler & Macro-F1 & Sampler & MCC \\
 
    \midrule
    \midrule
     AutoSMOTE & \textbf{8.83}$\:\:\:\;$ & AutoSMOTE & \textbf{15.08}$\:\;$ \\
     SupervisedSMOTE & 15.22$\blacktriangle$ & SupervisedSMOTE & 16.56$\:\:\;$ \\
     GASMOTE & 15.25$\blacktriangle$ & ClusterSMOTE & 27.29$\blacktriangle$ \\
     ClusterSMOTE & 18.58$\blacktriangle$ & ANDSMOTE & 28.39$\blacktriangle$ \\
     SMOTEPSO & 22.11$\blacktriangle$ & BorderlineSMOTE1 & 28.94$\blacktriangle$ \\

     \bottomrule
    \end{tabular}
    \vspace{-10pt}
\end{table}

\subsection{Comparison with SMOTE Variants}
\label{sec:43}
Since AutoSMOTE builds upon the SMOTE search space, we compare AutoSMOTE with the numerous SMOTE variants to investigate \textbf{RQ2}. We report the overall ranks of AutoSMOTE against 85 SMOTE variants across all the datasets in Table~\ref{tab:variants}. We observe that AutoSMOTE significantly outperforms the 85 SMOTE variants. It is ranked 8.85 and 15.08 among all the samplers in terms of Macro-F1 and MCC, respectively. We note that AutoSMOTE shares the same search space as the majority of the SMOTE variants. Thus, the improvement can be mainly attributed to searching with RL.

\subsection{Ablation Study}
\label{sec:44}

To study \textbf{RQ3}, we compare AutoSMOTE with several ablations. First, we remove each of the three policies in the hierarchy to show that each level of the policies contributes to the performance. Specifically, for removing the cross-instance sub-policy, we make $g_1 = G_1 / 2$ (which is the mean value of $g_1$ when we use cross-instance sub-policy). For removing the instance-specific sub-policy, we assume $g_2 = G_2 / 2$. For removing the low-level policy, we assume the neighbors and the interpolation weights are randomly selected. Second, we consider a flat policy baseline, which directly makes the low-level decisions. Specifically, the flat policy directly predicts the neighbors, interpolation weights, and a boolean indicating whether to go to the next instance. The flat policy is also trained with RL. Third, we consider a random search baseline\footnote{We have tried applying the existing techniques designed for hyperparameter tuning or neural architecture search to our problem. However, we find they are often not applicable because they cannot deal with the hierarchical search space, where the low-level search space has variable sizes since it depends on the high-level decisions.}, which randomly makes the high-level and low-level decisions within the same decision space as AutoSMOTE. The generated synthetic samples that lead to the best validation performance are used for evaluation. Fourth, we consider a variant that trains the classifier on both the training set and validation set. Specifically, we merge the original training and validation sets to form a new training set. Then the classifier and AutoSMOTE are both trained on the new training set. This ablation is designed to study whether it is necessary to separate a validation set. For a fair comparison, we set $I=1000$ for AutoSMOTE and all the ablations.

\begin{table}[t]
    \centering
    \footnotesize
    \caption{Average ranks of AutoSMOTE and the ablations on Mozilla4. $\blacktriangle$ suggests full AutoSMOTE is significantly better.}
    \vspace{-10pt}
    \label{tab:ablations}
    \setlength{\tabcolsep}{15pt}
    \begin{tabular}{l|cc}
    \toprule
     
  & Macro-F1 & MCC \\
 
    \midrule
    \midrule
     w/o cross-instance sub-policy & 3.79$\blacktriangle$ & 4.12$\blacktriangle$ \\
     w/o instance-specific sub-policy  & 2.75$\:\:\;$ & 3.75$\blacktriangle$ \\
     w/o low-level policy & 4.58$\blacktriangle$ & 4.21$\blacktriangle$ \\
     Flat policy & 4.67$\blacktriangle$ & 5.25$\blacktriangle$ \\
     Random search & 6.92$\blacktriangle$ & 5.75$\blacktriangle$ \\
     Merging training and validation sets & 2.83$\:\;$ & 2.83$\:\;$ \\
     \midrule
     Full AutoSMOTE & \textbf{2.46}$\:\;$ & \textbf{2.08}$\:\;$ \\

     \bottomrule
    \end{tabular}
    \vspace{-10pt}
\end{table}

Table~\ref{tab:ablations} shows the ranks of AutoSMOTE and the ablations on the Mozilla4 dataset. We observe that AutoSMOTE outperforms all the ablations. First, removing either of the three policies will degrade the performance, which demonstrates the necessity of modeling each level of the decision. Second, the flat policy is worse than AutoSMOTE, which is expected because the flat policy suffers from very long MDP, making RL harder to train. Third, random search shows very poor performance. This is because it is hard to identify a strong synthetic strategy within the massive search space of SMOTE, which demonstrates the effectiveness of RL. Finally, although merging training and validation sets can provide more data to the classifier, it may negatively affect the performance, which could be explained by over-fitting. We observe that it can achieve near-perfect performance on the training set while the testing performance remains low. Thus, it is necessary to separate a validation set so that we can optimize the generalization performance.

\begin{figure}[t]
  \centering
  \begin{subfigure}[b]{0.45\textwidth}
    \centering
    \includegraphics[width=0.99\textwidth]{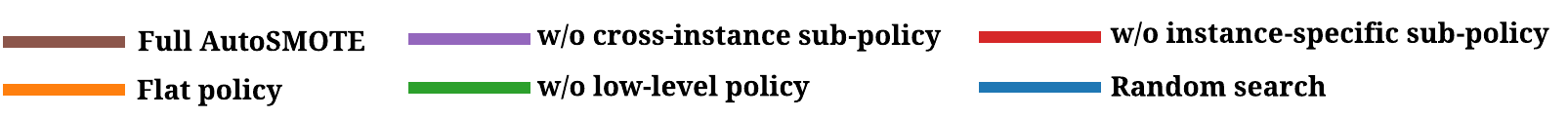}
  \end{subfigure}%

  \begin{subfigure}[b]{0.2\textwidth}
    \centering
    \includegraphics[width=0.99\textwidth]{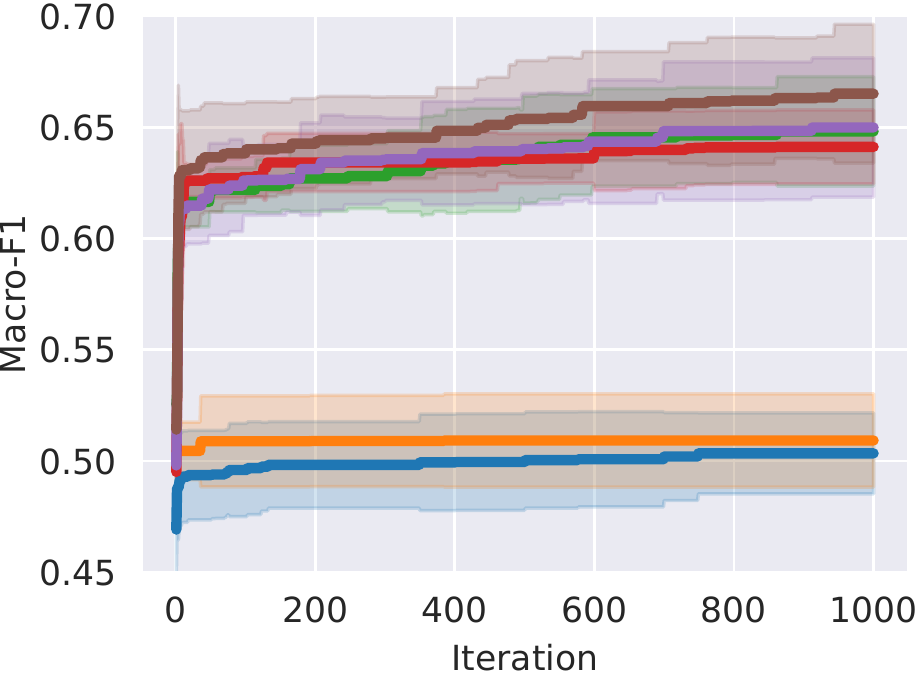}
  \end{subfigure}%
  \begin{subfigure}[b]{0.2\textwidth}
    \centering
    \includegraphics[width=0.99\textwidth]{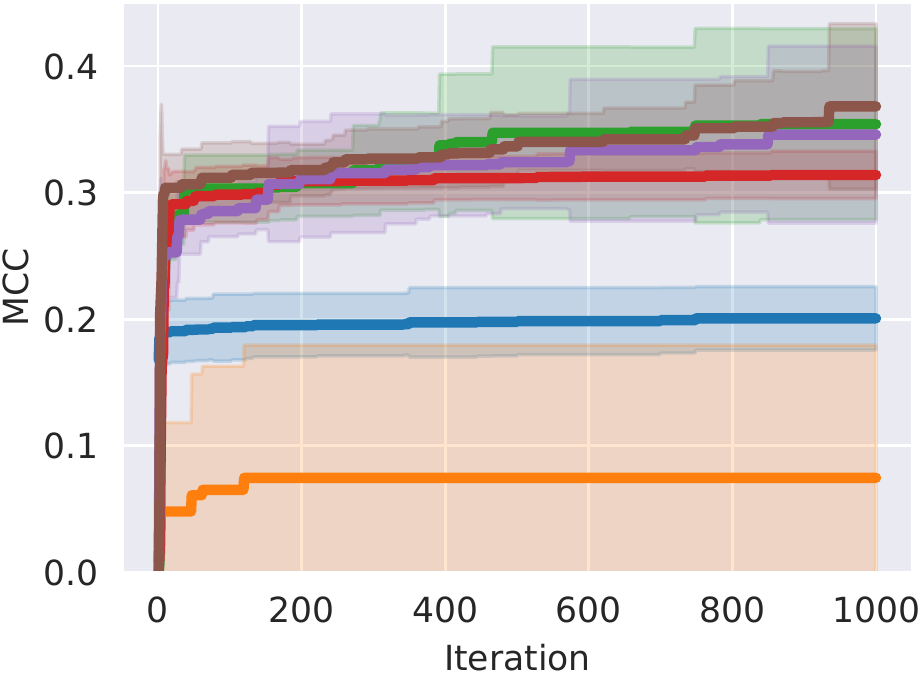}
  \end{subfigure}%
  \vspace{-10pt}
  \caption{Performance of AutoSMOTE and the ablations w.r.t. the number of searching iterations on Mozilla4 with imbalanced ratio of 50 and base classifier of SVM.}
  \vspace{-10pt}
  \label{fig:ablationsplot}
\end{figure}

To better understand the searching efficiency, we visualize in Figure~\ref{fig:ablationsplot} the best validation performance w.r.t. the number of search iterations on Mozilla4 with an imbalanced ratio of 50 and SVM as the base classifier. Note that we have excluded the ablation that merges the validation set because it tends to over-fit the validation data so that its validation performance is meaningless. We can observe that AutoSMOTE achieves better results in terms of both final performance and sample efficiency, i.e., achieving good performance with less number of iterations. We also observe that the flat policy and random search get stuck during the search, which again verifies the superiority of hierarchical RL.


\subsection{Analysis of the Hyperparameters}
\label{sec:45}

\begin{figure}[t]
  \centering

  \begin{subfigure}[b]{0.15\textwidth}
    \centering
    \includegraphics[width=0.99\textwidth]{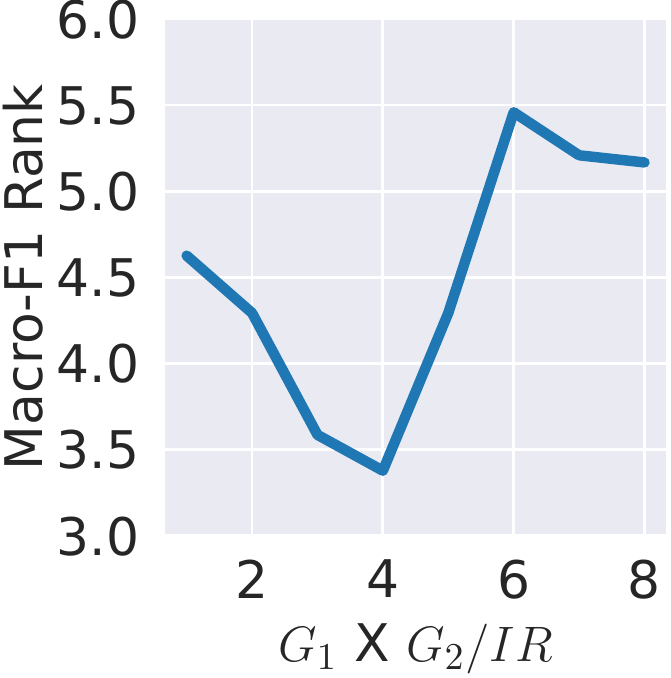}
  \end{subfigure}%
  \begin{subfigure}[b]{0.15\textwidth}
    \centering
    \includegraphics[width=0.99\textwidth]{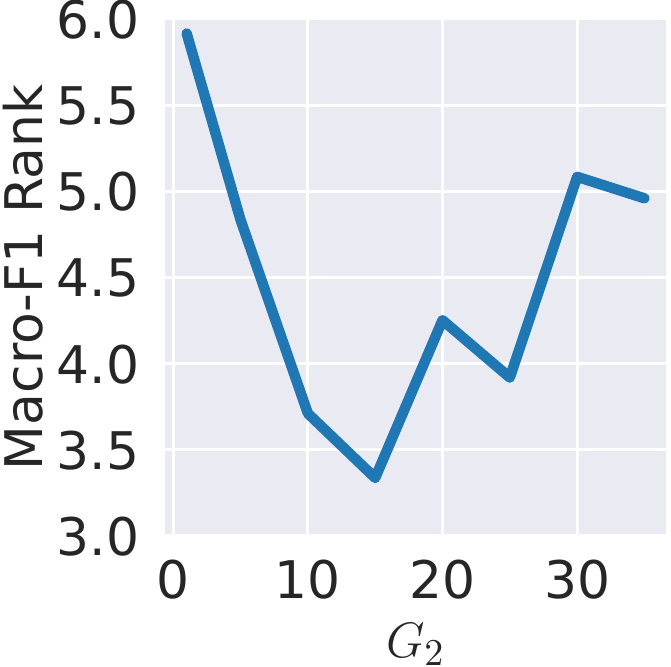}
  \end{subfigure}%
  \begin{subfigure}[b]{0.15\textwidth}
    \centering
    \includegraphics[width=0.99\textwidth]{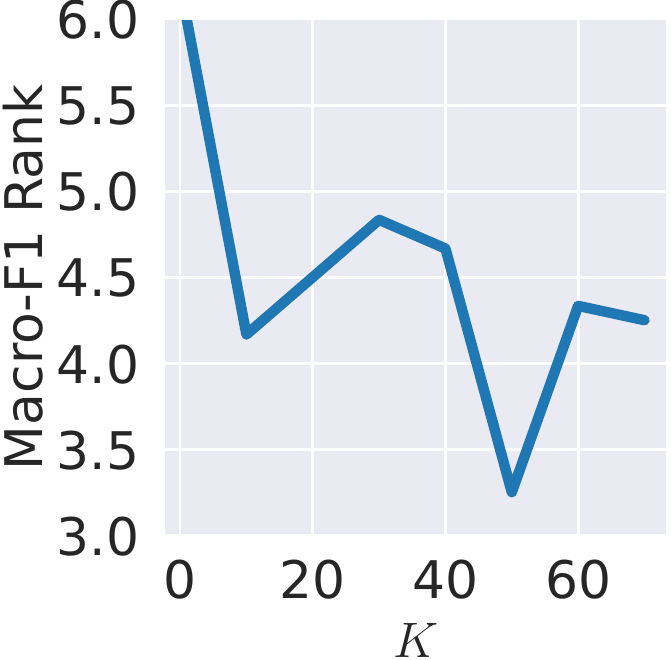}
  \end{subfigure}%
  \vspace{-5pt}
  \caption{Hyperparameter study on Mozilla4.}
  \vspace{-9pt}
  \label{fig:hyper}
\end{figure}

For \textbf{RQ4}, we study the impact of $G_1$, $G_2$, and $K$ on Mozilla4 dataset. First, we fix $G_2=10$ and $K=30$, and vary $G_1$ such that $G_1 \times G_2 / \text{IR}$ ranges from 1 to 8 (left-hand side of Figure~\ref{fig:hyper}). The best performance is achieved when $G_1 \times G_2 / \text{IR}=4$. A possible explanation is that a too low ratio will make the search space too restricted to discover good synthetic strategies, while a too-large ratio will make searching more difficult. Second, we fix $G_1 \times G_2 / \text{IR}=4 $ and $K=30$, and vary $G_2$ (middle of Figure~\ref{fig:hyper}). We find a too small $G_2$ will worsen the performance. Recall that the instance-specific sub-policy makes personalized decisions. A small $G_2$ will restrict the personalization, which could explain why it causes unsatisfactory performance. Similarly, we observe a performance drop when $G_2$ is large, which could also be explained by the difficulty brought by the larger search space. Third, we fix $G_1$ and $G_2$, and vary $K$. We observe a significant performance drop when $K$ is very small, which verifies the effectiveness of performing interpolation. 

Overall, we observe that $G_1$, $G_2$, and $K$ will control the trade-off between performance and searching efficiency in each level of the decisions. If the value is too small, it may restrict the search space and lead to worse performance. In contrast, a too-large value tends to make the searching more difficult and may also negatively affect the performance given a limited searching budget.

\subsection{Case Study}
\label{sec:46}

\begin{figure}[t]
  \centering
  \begin{subfigure}[b]{0.20\textwidth}
    \centering
    \includegraphics[width=0.99\textwidth]{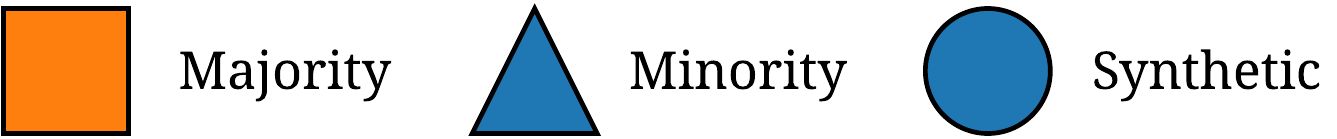}
  \end{subfigure}%

  \begin{subfigure}[b]{0.15\textwidth}
    \centering
    \includegraphics[width=0.99\textwidth]{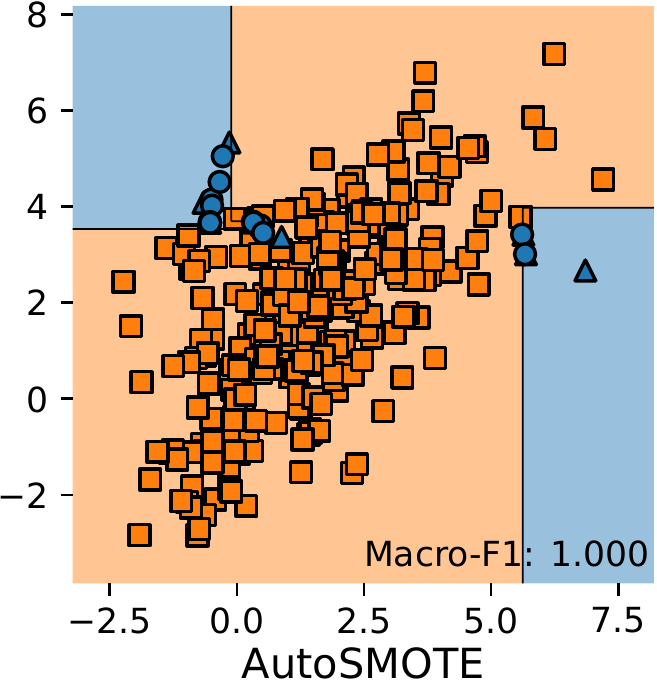}
  \end{subfigure}%
  \begin{subfigure}[b]{0.15\textwidth}
    \centering
    \includegraphics[width=0.99\textwidth]{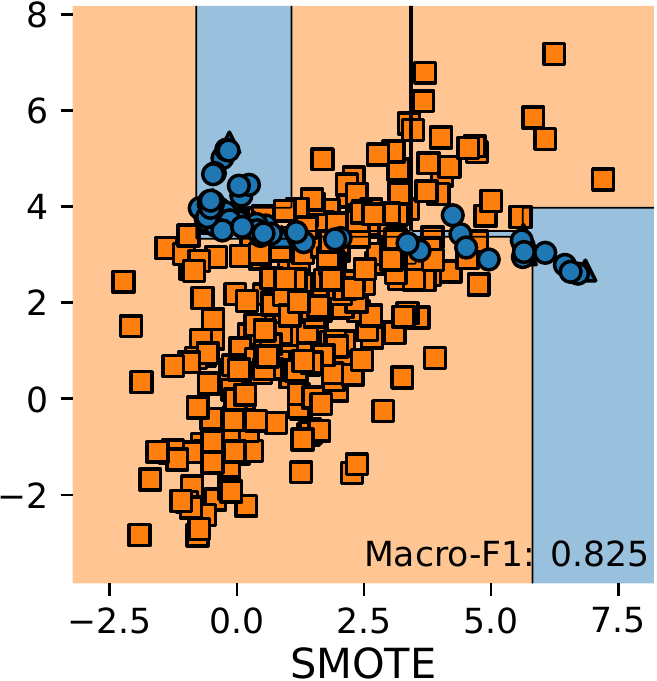}
  \end{subfigure}%
  \begin{subfigure}[b]{0.15\textwidth}
    \centering
    \includegraphics[width=0.99\textwidth]{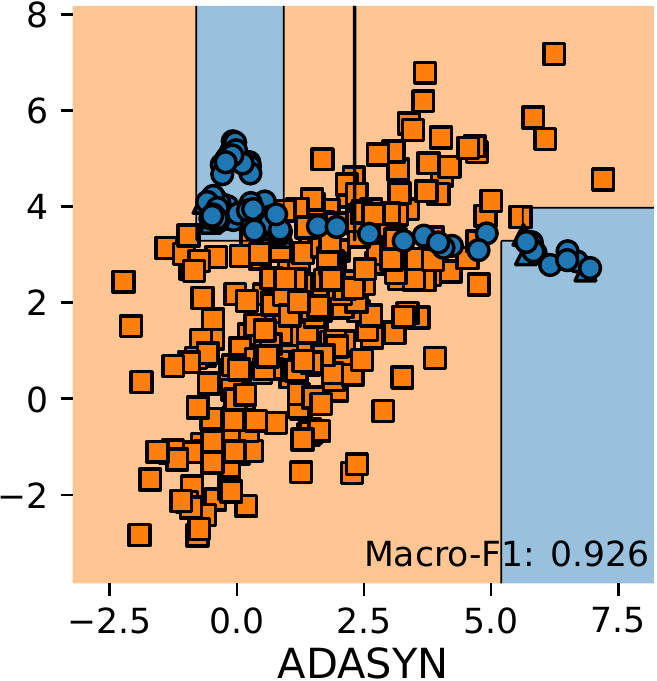}
  \end{subfigure}%
  
  \begin{subfigure}[b]{0.15\textwidth}
    \centering
    \includegraphics[width=0.99\textwidth]{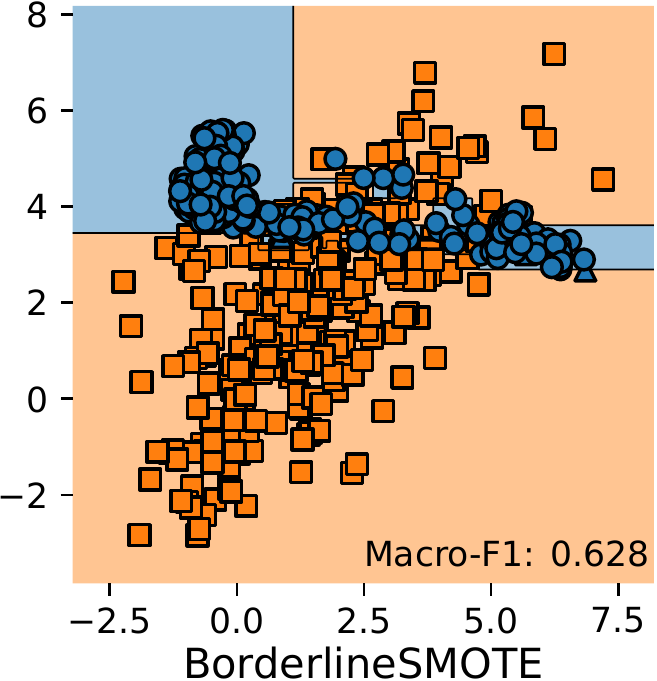}
  \end{subfigure}%
  \begin{subfigure}[b]{0.15\textwidth}
    \centering
    \includegraphics[width=0.99\textwidth]{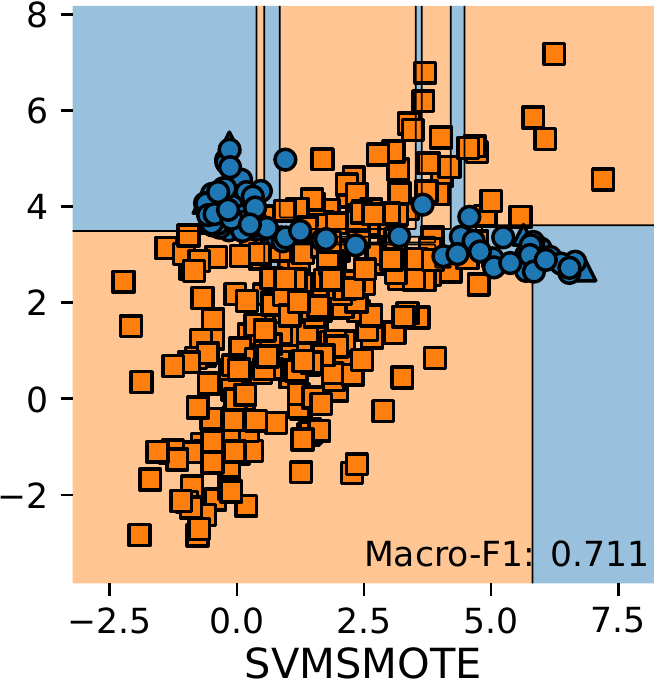}
  \end{subfigure}%
  \begin{subfigure}[b]{0.15\textwidth}
    \centering
    \includegraphics[width=0.99\textwidth]{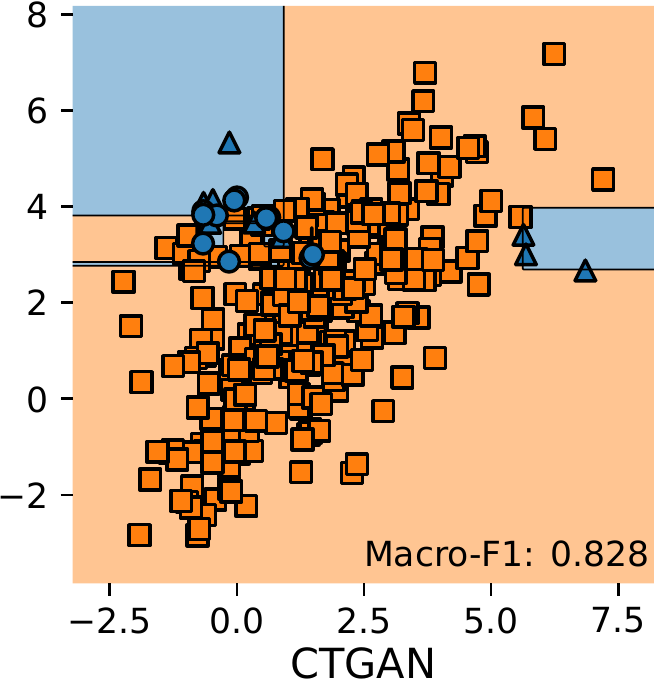}
  \end{subfigure}%
  \vspace{-5pt}
  \caption{Visualization of the generated synthetic samples and the decision boundary of DecisionTree  on a toy data using AutoSMOTE and other over-sampling techniques.}
  \vspace{-10pt}
  \label{fig:case}
\end{figure}

To answer \textbf{RQ5}, we apply AutoSMOTE, several SMOTE variants, and CTGAN to a 2-dimensional toy dataset. This dataset is publicly available\footnote{\url{https://github.com/shubhomoydas/ad_examples/tree/master/ad_examples/datasets/anomaly/toy2/fullsamples}} and is originally synthesized for anomaly detection. We under-sample the minority instances with an imbalanced ratio of 30, which results in 450 majority instances and 35 minority instances. Then we split 60\% of the data for training, 20\% of the data for validation, and 20\% for testing. This toy dataset is challenging because the minorities form two clusters in the two sides of the majority instances so that the sampler could easily generate noisy samples. We use Macro-F1 as the performance metric. Figure~\ref{fig:case} illustrates the generated synthetic samples and the decision boundary obtained by training a DecisionTree classifier on the over-sampled data. We observe that all the samplers except AutoSMOTE tend to generate some noisy samples that interleave with the majorities between the two clusters, which degrades the performance. In contrast, AutoSMOTE can identify a tailored synthetic strategy that achieves 1.0 Macro-F1 through directly optimizing the performance metric, which demonstrates the effectiveness of the learning-based sampler.


\section{Related Work}

\textbf{Imbalanced Learning.} Existing methods can be mainly grouped into three categories: data-level methods that aim to balance the data distributions by modifying the training data~\cite{wang2018towards,rout2018handling,johnson2019survey,wang2020global,zhao2021graphsmote,cai2019supervised}, algorithm-level approaches that try to modify the classifier such as the loss function~\cite{liu2006influence,li2021autobalance}, and hybrid solutions that combine the above two~\cite{krawczyk2016learning}. Our work falls into data-level methods. Data-level methods can be further divided into three sub-categories, including generating new samples for the minority class (over-sampling)~\cite{chawla2002smote,he2008adasyn}, removing instances from the majority class (under-sampling)~\cite{yen2006under}, and hybrid methods that combine the above two~\cite{more2016survey}. One advantage of over-sampling is that it will not lose information. However, it may be prone to over-fitting. AutoSMOTE tackles this problem by optimizing the generalization performance, thereby alleviating the over-fitting issue. A recent work~\cite{liu2020mesa} proposes a meta-learning algorithm, which similarly aims to optimize the performance metric on the validation data. However, they focus on learning ensemble strategies for under-sampling, which may lose information. In contrast, AutoSMOTE is an over-sampling method so that the resampled data can preserve all the information. Another line of imbalanced learning work is outlier detection~\cite{han2022adbench,lai2021revisiting}, which is often an unsupervised task.

\textbf{Automated Machine Learning (AutoML).} AutoML techniques have recently show promise in various data mining tasks~\cite{li2020autost,yang2020automl,koch2018autotune,liu2019automating,yang2019oboe,zhaok2021autoemb,zha2021autovideo,li2021automated,lai2021tods,zha2020meta,wang2022auto,li2021autood,zhao2021automatic,li2020pyodds,zha2022autoshard}. The key idea is to leverage machine learning to optimize data mining solutions, such as neural architecture search~\cite{elsken2019neural}, hyperparameter tuning~\cite{yu2020hyper}, and pipeline search~\cite{heffetz2020deepline}. AutoSMOTE also falls into this line of research. In contrast to the previous work, over-sampling exhibits unique challenges with a huge and complex decision space that requires hierarchical reasoning. The existing AutoML approaches often can only deal with simple and flat search spaces. To this end, we develop a tailored search space based on SMOTE and leverage deep hierarchical RL to jointly optimize different levels of the decisions.

\textbf{Deep RL.} Deep RL has achieved remarkable success in games~\cite{mnih2013playing,zha2021rlcard,zha2021douzero,espeholt2018impala,zha2019experience,zha2021rank,zha2021simplifying}. Deep RL is designed for goal-oriented tasks, where the agent is trained based on the reward signal. Recently, deep hierarchical RL has shown promise in tackling tasks with long horizons~\cite{pateria2021hierarchical}. The key idea is decompose the MDP into decisions in different granularities such that each level of the decisions can be more effectively leaned. However, the successes of RL are often only demonstrated in simulated games. Our work suggests that deep hierarchical RL can help tackle the data imbalance problem, which is a real-world data mining challenge.

\section{Conclusions and Future Work}

This work investigates learning-based sampling algorithms for tackling the imbalanced learning problem. To this end, we first formulate a hierarchical search space based on SMOTE. Then we leverage deep hierarchical RL to jointly optimize different levels of decisions to achieve the best generalization performance on the validation set. The proposed over-sampling algorithm, namely AutoSMOTE, is evaluated against the state-of-the-art samplers and also numerous SMOTE variants. Extensive experiments demonstrate the superiority of AutoSMOTE over heuristics. In the future, we will extend AutoSMOTE to perform combined over- and under-sampling and deal with other data types such as images, graphs, and time series.

\section*{Acknowledgements}
The work is, in part, supported by NSF (\#IIS-1849085, \#IIS-1900990, \#IIS-1939716). The views and conclusions in this paper should not be interpreted as representing any funding agencies.

\newpage

\bibliographystyle{ACM-Reference-Format}
\balance
\bibliography{ref}


\begin{thebibliography}{56}


\ifx \showCODEN    \undefined \def \showCODEN     #1{\unskip}     \fi
\ifx \showDOI      \undefined \def \showDOI       #1{#1}\fi
\ifx \showISBNx    \undefined \def \showISBNx     #1{\unskip}     \fi
\ifx \showISBNxiii \undefined \def \showISBNxiii  #1{\unskip}     \fi
\ifx \showISSN     \undefined \def \showISSN      #1{\unskip}     \fi
\ifx \showLCCN     \undefined \def \showLCCN      #1{\unskip}     \fi
\ifx \shownote     \undefined \def \shownote      #1{#1}          \fi
\ifx \showarticletitle \undefined \def \showarticletitle #1{#1}   \fi
\ifx \showURL      \undefined \def \showURL       {\relax}        \fi
\providecommand\bibfield[2]{#2}
\providecommand\bibinfo[2]{#2}
\providecommand\natexlab[1]{#1}
\providecommand\showeprint[2][]{arXiv:#2}

\bibitem[\protect\citeauthoryear{Boughorbel, Jarray, and El-Anbari}{Boughorbel
  et~al\mbox{.}}{2017}]%
        {boughorbel2017optimal}
\bibfield{author}{\bibinfo{person}{Sabri Boughorbel}, \bibinfo{person}{Fethi
  Jarray}, {and} \bibinfo{person}{Mohammed El-Anbari}.}
  \bibinfo{year}{2017}\natexlab{}.
\newblock \showarticletitle{Optimal classifier for imbalanced data using
  Matthews Correlation Coefficient metric}.
\newblock \bibinfo{journal}{\emph{PloS one}} \bibinfo{volume}{12},
  \bibinfo{number}{6} (\bibinfo{year}{2017}), \bibinfo{pages}{e0177678}.
\newblock


\bibitem[\protect\citeauthoryear{Buda, Maki, and Mazurowski}{Buda
  et~al\mbox{.}}{2018}]%
        {buda2018systematic}
\bibfield{author}{\bibinfo{person}{Mateusz Buda}, \bibinfo{person}{Atsuto
  Maki}, {and} \bibinfo{person}{Maciej~A Mazurowski}.}
  \bibinfo{year}{2018}\natexlab{}.
\newblock \showarticletitle{A systematic study of the class imbalance problem
  in convolutional neural networks}.
\newblock \bibinfo{journal}{\emph{Neural Networks}}  \bibinfo{volume}{106}
  (\bibinfo{year}{2018}), \bibinfo{pages}{249--259}.
\newblock


\bibitem[\protect\citeauthoryear{Cai, Wang, Zhou, Xu, and Jing}{Cai
  et~al\mbox{.}}{2019}]%
        {cai2019supervised}
\bibfield{author}{\bibinfo{person}{Zixin Cai}, \bibinfo{person}{Xinyue Wang},
  \bibinfo{person}{Mingjie Zhou}, \bibinfo{person}{Jian Xu}, {and}
  \bibinfo{person}{Liping Jing}.} \bibinfo{year}{2019}\natexlab{}.
\newblock \showarticletitle{Supervised class distribution learning for
  GANs-based imbalanced classification}. In \bibinfo{booktitle}{\emph{ICDM}}.
\newblock


\bibitem[\protect\citeauthoryear{Chawla, Bowyer, Hall, and Kegelmeyer}{Chawla
  et~al\mbox{.}}{2002}]%
        {chawla2002smote}
\bibfield{author}{\bibinfo{person}{Nitesh~V Chawla}, \bibinfo{person}{Kevin~W
  Bowyer}, \bibinfo{person}{Lawrence~O Hall}, {and} \bibinfo{person}{W~Philip
  Kegelmeyer}.} \bibinfo{year}{2002}\natexlab{}.
\newblock \showarticletitle{SMOTE: synthetic minority over-sampling technique}.
\newblock \bibinfo{journal}{\emph{Journal of artificial intelligence research}}
   \bibinfo{volume}{16} (\bibinfo{year}{2002}), \bibinfo{pages}{321--357}.
\newblock


\bibitem[\protect\citeauthoryear{Chawla, Japkowicz, and Kotcz}{Chawla
  et~al\mbox{.}}{2004}]%
        {chawla2004special}
\bibfield{author}{\bibinfo{person}{Nitesh~V Chawla}, \bibinfo{person}{Nathalie
  Japkowicz}, {and} \bibinfo{person}{Aleksander Kotcz}.}
  \bibinfo{year}{2004}\natexlab{}.
\newblock \showarticletitle{Special issue on learning from imbalanced data
  sets}.
\newblock \bibinfo{journal}{\emph{ACM SIGKDD explorations newsletter}}
  \bibinfo{volume}{6}, \bibinfo{number}{1} (\bibinfo{year}{2004}),
  \bibinfo{pages}{1--6}.
\newblock


\bibitem[\protect\citeauthoryear{Elsken, Metzen, and Hutter}{Elsken
  et~al\mbox{.}}{2019}]%
        {elsken2019neural}
\bibfield{author}{\bibinfo{person}{Thomas Elsken}, \bibinfo{person}{Jan~Hendrik
  Metzen}, {and} \bibinfo{person}{Frank Hutter}.}
  \bibinfo{year}{2019}\natexlab{}.
\newblock \showarticletitle{Neural architecture search: A survey}.
\newblock \bibinfo{journal}{\emph{The Journal of Machine Learning Research}}
  \bibinfo{volume}{20}, \bibinfo{number}{1} (\bibinfo{year}{2019}),
  \bibinfo{pages}{1997--2017}.
\newblock


\bibitem[\protect\citeauthoryear{Espeholt, Soyer, Munos, Simonyan, Mnih, Ward,
  Doron, Firoiu, Harley, Dunning, et~al\mbox{.}}{Espeholt
  et~al\mbox{.}}{2018}]%
        {espeholt2018impala}
\bibfield{author}{\bibinfo{person}{Lasse Espeholt}, \bibinfo{person}{Hubert
  Soyer}, \bibinfo{person}{Remi Munos}, \bibinfo{person}{Karen Simonyan},
  \bibinfo{person}{Vlad Mnih}, \bibinfo{person}{Tom Ward},
  \bibinfo{person}{Yotam Doron}, \bibinfo{person}{Vlad Firoiu},
  \bibinfo{person}{Tim Harley}, \bibinfo{person}{Iain Dunning},
  {et~al\mbox{.}}} \bibinfo{year}{2018}\natexlab{}.
\newblock \showarticletitle{Impala: Scalable distributed deep-rl with
  importance weighted actor-learner architectures}. In
  \bibinfo{booktitle}{\emph{ICML}}.
\newblock


\bibitem[\protect\citeauthoryear{Han, Wang, and Mao}{Han et~al\mbox{.}}{2005}]%
        {han2005borderline}
\bibfield{author}{\bibinfo{person}{Hui Han}, \bibinfo{person}{Wen-Yuan Wang},
  {and} \bibinfo{person}{Bing-Huan Mao}.} \bibinfo{year}{2005}\natexlab{}.
\newblock \showarticletitle{Borderline-SMOTE: a new over-sampling method in
  imbalanced data sets learning}. In \bibinfo{booktitle}{\emph{ICIC}}.
\newblock


\bibitem[\protect\citeauthoryear{Han, Hu, Huang, Jiang, and Zhao}{Han
  et~al\mbox{.}}{2022}]%
        {han2022adbench}
\bibfield{author}{\bibinfo{person}{Songqiao Han}, \bibinfo{person}{Xiyang Hu},
  \bibinfo{person}{Hailiang Huang}, \bibinfo{person}{Mingqi Jiang}, {and}
  \bibinfo{person}{Yue Zhao}.} \bibinfo{year}{2022}\natexlab{}.
\newblock \showarticletitle{ADBench: Anomaly Detection Benchmark}.
\newblock \bibinfo{journal}{\emph{arXiv preprint arXiv:2206.09426}}
  (\bibinfo{year}{2022}).
\newblock


\bibitem[\protect\citeauthoryear{He, Bai, Garcia, and Li}{He
  et~al\mbox{.}}{2008}]%
        {he2008adasyn}
\bibfield{author}{\bibinfo{person}{Haibo He}, \bibinfo{person}{Yang Bai},
  \bibinfo{person}{Edwardo~A Garcia}, {and} \bibinfo{person}{Shutao Li}.}
  \bibinfo{year}{2008}\natexlab{}.
\newblock \showarticletitle{ADASYN: Adaptive synthetic sampling approach for
  imbalanced learning}. In \bibinfo{booktitle}{\emph{IJCNN}}.
\newblock


\bibitem[\protect\citeauthoryear{Heffetz, Vainshtein, Katz, and Rokach}{Heffetz
  et~al\mbox{.}}{2020}]%
        {heffetz2020deepline}
\bibfield{author}{\bibinfo{person}{Yuval Heffetz}, \bibinfo{person}{Roman
  Vainshtein}, \bibinfo{person}{Gilad Katz}, {and} \bibinfo{person}{Lior
  Rokach}.} \bibinfo{year}{2020}\natexlab{}.
\newblock \showarticletitle{Deepline: Automl tool for pipelines generation
  using deep reinforcement learning and hierarchical actions filtering}. In
  \bibinfo{booktitle}{\emph{KDD}}.
\newblock


\bibitem[\protect\citeauthoryear{Johnson and Khoshgoftaar}{Johnson and
  Khoshgoftaar}{2019}]%
        {johnson2019survey}
\bibfield{author}{\bibinfo{person}{Justin~M Johnson} {and}
  \bibinfo{person}{Taghi~M Khoshgoftaar}.} \bibinfo{year}{2019}\natexlab{}.
\newblock \showarticletitle{Survey on deep learning with class imbalance}.
\newblock \bibinfo{journal}{\emph{Journal of Big Data}} \bibinfo{volume}{6},
  \bibinfo{number}{1} (\bibinfo{year}{2019}), \bibinfo{pages}{1--54}.
\newblock


\bibitem[\protect\citeauthoryear{Koch, Golovidov, Gardner, Wujek, Griffin, and
  Xu}{Koch et~al\mbox{.}}{2018}]%
        {koch2018autotune}
\bibfield{author}{\bibinfo{person}{Patrick Koch}, \bibinfo{person}{Oleg
  Golovidov}, \bibinfo{person}{Steven Gardner}, \bibinfo{person}{Brett Wujek},
  \bibinfo{person}{Joshua Griffin}, {and} \bibinfo{person}{Yan Xu}.}
  \bibinfo{year}{2018}\natexlab{}.
\newblock \showarticletitle{Autotune: A derivative-free optimization framework
  for hyperparameter tuning}. In \bibinfo{booktitle}{\emph{KDD}}.
\newblock


\bibitem[\protect\citeauthoryear{Kov{\'a}cs}{Kov{\'a}cs}{2019a}]%
        {kovacs2019empirical}
\bibfield{author}{\bibinfo{person}{Gy{\"o}rgy Kov{\'a}cs}.}
  \bibinfo{year}{2019}\natexlab{a}.
\newblock \showarticletitle{An empirical comparison and evaluation of minority
  oversampling techniques on a large number of imbalanced datasets}.
\newblock \bibinfo{journal}{\emph{Applied Soft Computing}}
  \bibinfo{volume}{83} (\bibinfo{year}{2019}), \bibinfo{pages}{105662}.
\newblock


\bibitem[\protect\citeauthoryear{Kov{\'a}cs}{Kov{\'a}cs}{2019b}]%
        {kovacs2019smote}
\bibfield{author}{\bibinfo{person}{Gy{\"o}rgy Kov{\'a}cs}.}
  \bibinfo{year}{2019}\natexlab{b}.
\newblock \showarticletitle{Smote-variants: A python implementation of 85
  minority oversampling techniques}.
\newblock \bibinfo{journal}{\emph{Neurocomputing}}  \bibinfo{volume}{366}
  (\bibinfo{year}{2019}), \bibinfo{pages}{352--354}.
\newblock


\bibitem[\protect\citeauthoryear{Krawczyk}{Krawczyk}{2016}]%
        {krawczyk2016learning}
\bibfield{author}{\bibinfo{person}{Bartosz Krawczyk}.}
  \bibinfo{year}{2016}\natexlab{}.
\newblock \showarticletitle{Learning from imbalanced data: open challenges and
  future directions}.
\newblock \bibinfo{journal}{\emph{Progress in Artificial Intelligence}}
  \bibinfo{volume}{5}, \bibinfo{number}{4} (\bibinfo{year}{2016}),
  \bibinfo{pages}{221--232}.
\newblock


\bibitem[\protect\citeauthoryear{Lai, Zha, Wang, Xu, Zhao, Kumar, Chen,
  Zumkhawaka, Wan, Martinez, et~al\mbox{.}}{Lai et~al\mbox{.}}{2021a}]%
        {lai2021tods}
\bibfield{author}{\bibinfo{person}{Kwei-Herng Lai}, \bibinfo{person}{Daochen
  Zha}, \bibinfo{person}{Guanchu Wang}, \bibinfo{person}{Junjie Xu},
  \bibinfo{person}{Yue Zhao}, \bibinfo{person}{Devesh Kumar},
  \bibinfo{person}{Yile Chen}, \bibinfo{person}{Purav Zumkhawaka},
  \bibinfo{person}{Minyang Wan}, \bibinfo{person}{Diego Martinez},
  {et~al\mbox{.}}} \bibinfo{year}{2021}\natexlab{a}.
\newblock \showarticletitle{TODS: An Automated Time Series Outlier Detection
  System}. In \bibinfo{booktitle}{\emph{AAAI}}.
\newblock


\bibitem[\protect\citeauthoryear{Lai, Zha, Xu, Zhao, Wang, and Hu}{Lai
  et~al\mbox{.}}{2021b}]%
        {lai2021revisiting}
\bibfield{author}{\bibinfo{person}{Kwei-Herng Lai}, \bibinfo{person}{Daochen
  Zha}, \bibinfo{person}{Junjie Xu}, \bibinfo{person}{Yue Zhao},
  \bibinfo{person}{Guanchu Wang}, {and} \bibinfo{person}{Xia Hu}.}
  \bibinfo{year}{2021}\natexlab{b}.
\newblock \showarticletitle{Revisiting time series outlier detection:
  Definitions and benchmarks}. In \bibinfo{booktitle}{\emph{NeurIPS}}.
\newblock


\bibitem[\protect\citeauthoryear{Li, Zhang, Thrampoulidis, Chen, and Oymak}{Li
  et~al\mbox{.}}{2021c}]%
        {li2021autobalance}
\bibfield{author}{\bibinfo{person}{Mingchen Li}, \bibinfo{person}{Xuechen
  Zhang}, \bibinfo{person}{Christos Thrampoulidis}, \bibinfo{person}{Jiasi
  Chen}, {and} \bibinfo{person}{Samet Oymak}.}
  \bibinfo{year}{2021}\natexlab{c}.
\newblock \showarticletitle{AutoBalance: Optimized Loss Functions for
  Imbalanced Data}.
\newblock \bibinfo{journal}{\emph{NeurIPS}} (\bibinfo{year}{2021}).
\newblock


\bibitem[\protect\citeauthoryear{Li, Zhang, Bao, Liang, Li, and Zheng}{Li
  et~al\mbox{.}}{2020b}]%
        {li2020autost}
\bibfield{author}{\bibinfo{person}{Ting Li}, \bibinfo{person}{Junbo Zhang},
  \bibinfo{person}{Kainan Bao}, \bibinfo{person}{Yuxuan Liang},
  \bibinfo{person}{Yexin Li}, {and} \bibinfo{person}{Yu Zheng}.}
  \bibinfo{year}{2020}\natexlab{b}.
\newblock \showarticletitle{Autost: Efficient neural architecture search for
  spatio-temporal prediction}. In \bibinfo{booktitle}{\emph{KDD}}.
\newblock


\bibitem[\protect\citeauthoryear{Li, Chen, Zha, Zhou, Jin, Chen, and Hu}{Li
  et~al\mbox{.}}{2021a}]%
        {li2021automated}
\bibfield{author}{\bibinfo{person}{Yuening Li}, \bibinfo{person}{Zhengzhang
  Chen}, \bibinfo{person}{Daochen Zha}, \bibinfo{person}{Kaixiong Zhou},
  \bibinfo{person}{Haifeng Jin}, \bibinfo{person}{Haifeng Chen}, {and}
  \bibinfo{person}{Xia Hu}.} \bibinfo{year}{2021}\natexlab{a}.
\newblock \showarticletitle{Automated Anomaly Detection via Curiosity-Guided
  Search and Self-Imitation Learning}.
\newblock \bibinfo{journal}{\emph{IEEE Transactions on Neural Networks and
  Learning Systems}} (\bibinfo{year}{2021}).
\newblock


\bibitem[\protect\citeauthoryear{Li, Chen, Zha, Zhou, Jin, Chen, and Hu}{Li
  et~al\mbox{.}}{2021b}]%
        {li2021autood}
\bibfield{author}{\bibinfo{person}{Yuening Li}, \bibinfo{person}{Zhengzhang
  Chen}, \bibinfo{person}{Daochen Zha}, \bibinfo{person}{Kaixiong Zhou},
  \bibinfo{person}{Haifeng Jin}, \bibinfo{person}{Haifeng Chen}, {and}
  \bibinfo{person}{Xia Hu}.} \bibinfo{year}{2021}\natexlab{b}.
\newblock \showarticletitle{Autood: Neural architecture search for outlier
  detection}. In \bibinfo{booktitle}{\emph{ICDE}}.
\newblock


\bibitem[\protect\citeauthoryear{Li, Zha, Venugopal, Zou, and Hu}{Li
  et~al\mbox{.}}{2020a}]%
        {li2020pyodds}
\bibfield{author}{\bibinfo{person}{Yuening Li}, \bibinfo{person}{Daochen Zha},
  \bibinfo{person}{Praveen Venugopal}, \bibinfo{person}{Na Zou}, {and}
  \bibinfo{person}{Xia Hu}.} \bibinfo{year}{2020}\natexlab{a}.
\newblock \showarticletitle{Pyodds: An end-to-end outlier detection system with
  automated machine learning}. In \bibinfo{booktitle}{\emph{WWW}}.
\newblock


\bibitem[\protect\citeauthoryear{Liu, Fu, Wang, Wu, Bo, and Li}{Liu
  et~al\mbox{.}}{2019}]%
        {liu2019automating}
\bibfield{author}{\bibinfo{person}{Kunpeng Liu}, \bibinfo{person}{Yanjie Fu},
  \bibinfo{person}{Pengfei Wang}, \bibinfo{person}{Le Wu}, \bibinfo{person}{Rui
  Bo}, {and} \bibinfo{person}{Xiaolin Li}.} \bibinfo{year}{2019}\natexlab{}.
\newblock \showarticletitle{Automating feature subspace exploration via
  multi-agent reinforcement learning}. In \bibinfo{booktitle}{\emph{KDD}}.
\newblock


\bibitem[\protect\citeauthoryear{Liu and Zhou}{Liu and Zhou}{2006}]%
        {liu2006influence}
\bibfield{author}{\bibinfo{person}{Xu-Ying Liu} {and} \bibinfo{person}{Zhi-Hua
  Zhou}.} \bibinfo{year}{2006}\natexlab{}.
\newblock \showarticletitle{The influence of class imbalance on cost-sensitive
  learning: An empirical study}. In \bibinfo{booktitle}{\emph{ICDM}}.
\newblock


\bibitem[\protect\citeauthoryear{Liu, Wei, Jiang, Cao, Bian, and Chang}{Liu
  et~al\mbox{.}}{2020}]%
        {liu2020mesa}
\bibfield{author}{\bibinfo{person}{Zhining Liu}, \bibinfo{person}{Pengfei Wei},
  \bibinfo{person}{Jing Jiang}, \bibinfo{person}{Wei Cao},
  \bibinfo{person}{Jiang Bian}, {and} \bibinfo{person}{Yi Chang}.}
  \bibinfo{year}{2020}\natexlab{}.
\newblock \showarticletitle{MESA: Boost Ensemble Imbalanced Learning with
  MEta-SAmpler}. In \bibinfo{booktitle}{\emph{NeurIPS}}.
\newblock


\bibitem[\protect\citeauthoryear{Mnih, Kavukcuoglu, Silver, Graves, Antonoglou,
  Wierstra, and Riedmiller}{Mnih et~al\mbox{.}}{2013}]%
        {mnih2013playing}
\bibfield{author}{\bibinfo{person}{Volodymyr Mnih}, \bibinfo{person}{Koray
  Kavukcuoglu}, \bibinfo{person}{David Silver}, \bibinfo{person}{Alex Graves},
  \bibinfo{person}{Ioannis Antonoglou}, \bibinfo{person}{Daan Wierstra}, {and}
  \bibinfo{person}{Martin Riedmiller}.} \bibinfo{year}{2013}\natexlab{}.
\newblock \showarticletitle{Playing atari with deep reinforcement learning}.
\newblock \bibinfo{journal}{\emph{arXiv preprint arXiv:1312.5602}}
  (\bibinfo{year}{2013}).
\newblock


\bibitem[\protect\citeauthoryear{More}{More}{2016}]%
        {more2016survey}
\bibfield{author}{\bibinfo{person}{Ajinkya More}.}
  \bibinfo{year}{2016}\natexlab{}.
\newblock \showarticletitle{Survey of resampling techniques for improving
  classification performance in unbalanced datasets}.
\newblock \bibinfo{journal}{\emph{arXiv preprint arXiv:1608.06048}}
  (\bibinfo{year}{2016}).
\newblock


\bibitem[\protect\citeauthoryear{Nguyen, Cooper, and Kamei}{Nguyen
  et~al\mbox{.}}{2011}]%
        {nguyen2011borderline}
\bibfield{author}{\bibinfo{person}{Hien~M Nguyen}, \bibinfo{person}{Eric~W
  Cooper}, {and} \bibinfo{person}{Katsuari Kamei}.}
  \bibinfo{year}{2011}\natexlab{}.
\newblock \showarticletitle{Borderline over-sampling for imbalanced data
  classification}.
\newblock \bibinfo{journal}{\emph{International Journal of Knowledge
  Engineering and Soft Data Paradigms}} \bibinfo{volume}{3},
  \bibinfo{number}{1} (\bibinfo{year}{2011}), \bibinfo{pages}{4--21}.
\newblock


\bibitem[\protect\citeauthoryear{Pateria, Subagdja, Tan, and Quek}{Pateria
  et~al\mbox{.}}{2021}]%
        {pateria2021hierarchical}
\bibfield{author}{\bibinfo{person}{Shubham Pateria}, \bibinfo{person}{Budhitama
  Subagdja}, \bibinfo{person}{Ah-hwee Tan}, {and} \bibinfo{person}{Chai Quek}.}
  \bibinfo{year}{2021}\natexlab{}.
\newblock \showarticletitle{Hierarchical Reinforcement Learning: A
  Comprehensive Survey}.
\newblock \bibinfo{journal}{\emph{ACM Computing Surveys (CSUR)}}
  \bibinfo{volume}{54}, \bibinfo{number}{5} (\bibinfo{year}{2021}),
  \bibinfo{pages}{1--35}.
\newblock


\bibitem[\protect\citeauthoryear{Rout, Mishra, and Mallick}{Rout
  et~al\mbox{.}}{2018}]%
        {rout2018handling}
\bibfield{author}{\bibinfo{person}{Neelam Rout}, \bibinfo{person}{Debahuti
  Mishra}, {and} \bibinfo{person}{Manas~Kumar Mallick}.}
  \bibinfo{year}{2018}\natexlab{}.
\newblock \showarticletitle{Handling imbalanced data: a survey}. In
  \bibinfo{booktitle}{\emph{ASISA}}.
\newblock


\bibitem[\protect\citeauthoryear{Santoso, Wijayanto, Notodiputro, and
  Sartono}{Santoso et~al\mbox{.}}{2017}]%
        {santoso2017synthetic}
\bibfield{author}{\bibinfo{person}{B Santoso}, \bibinfo{person}{H Wijayanto},
  \bibinfo{person}{KA Notodiputro}, {and} \bibinfo{person}{B Sartono}.}
  \bibinfo{year}{2017}\natexlab{}.
\newblock \showarticletitle{Synthetic over sampling methods for handling class
  imbalanced problems: A review}. In \bibinfo{booktitle}{\emph{IOP conference
  series: earth and environmental science}}, Vol.~\bibinfo{volume}{58}. IOP
  Publishing, \bibinfo{pages}{012031}.
\newblock


\bibitem[\protect\citeauthoryear{Shu, Sliva, Wang, Tang, and Liu}{Shu
  et~al\mbox{.}}{2017}]%
        {shu2017fake}
\bibfield{author}{\bibinfo{person}{Kai Shu}, \bibinfo{person}{Amy Sliva},
  \bibinfo{person}{Suhang Wang}, \bibinfo{person}{Jiliang Tang}, {and}
  \bibinfo{person}{Huan Liu}.} \bibinfo{year}{2017}\natexlab{}.
\newblock \showarticletitle{Fake news detection on social media: A data mining
  perspective}.
\newblock \bibinfo{journal}{\emph{ACM SIGKDD explorations newsletter}}
  \bibinfo{volume}{19}, \bibinfo{number}{1} (\bibinfo{year}{2017}),
  \bibinfo{pages}{22--36}.
\newblock


\bibitem[\protect\citeauthoryear{Siriseriwan and Sinapiromsaran}{Siriseriwan
  and Sinapiromsaran}{2017}]%
        {siriseriwan2017adaptive}
\bibfield{author}{\bibinfo{person}{Wacharasak Siriseriwan} {and}
  \bibinfo{person}{Krung Sinapiromsaran}.} \bibinfo{year}{2017}\natexlab{}.
\newblock \showarticletitle{Adaptive neighbor synthetic minority oversampling
  technique under 1NN outcast handling}.
\newblock \bibinfo{journal}{\emph{Songklanakarin J. Sci. Technol}}
  \bibinfo{volume}{39}, \bibinfo{number}{5} (\bibinfo{year}{2017}),
  \bibinfo{pages}{565--576}.
\newblock


\bibitem[\protect\citeauthoryear{Sutton and Barto}{Sutton and Barto}{2018}]%
        {sutton2018reinforcement}
\bibfield{author}{\bibinfo{person}{Richard~S Sutton} {and}
  \bibinfo{person}{Andrew~G Barto}.} \bibinfo{year}{2018}\natexlab{}.
\newblock \bibinfo{booktitle}{\emph{Reinforcement learning: An introduction}}.
\newblock \bibinfo{publisher}{MIT press}.
\newblock


\bibitem[\protect\citeauthoryear{Vanschoren, Van~Rijn, Bischl, and
  Torgo}{Vanschoren et~al\mbox{.}}{2014}]%
        {vanschoren2014openml}
\bibfield{author}{\bibinfo{person}{Joaquin Vanschoren}, \bibinfo{person}{Jan~N
  Van~Rijn}, \bibinfo{person}{Bernd Bischl}, {and} \bibinfo{person}{Luis
  Torgo}.} \bibinfo{year}{2014}\natexlab{}.
\newblock \showarticletitle{OpenML: networked science in machine learning}.
\newblock \bibinfo{journal}{\emph{ACM SIGKDD Explorations Newsletter}}
  \bibinfo{volume}{15}, \bibinfo{number}{2} (\bibinfo{year}{2014}),
  \bibinfo{pages}{49--60}.
\newblock


\bibitem[\protect\citeauthoryear{Wang and Zhang}{Wang and Zhang}{2018}]%
        {wang2018towards}
\bibfield{author}{\bibinfo{person}{Jing Wang} {and} \bibinfo{person}{Min-Ling
  Zhang}.} \bibinfo{year}{2018}\natexlab{}.
\newblock \showarticletitle{Towards mitigating the class-imbalance problem for
  partial label learning}. In \bibinfo{booktitle}{\emph{KDD}}.
\newblock


\bibitem[\protect\citeauthoryear{Wang, Wang, Fan, Liu, and Tang}{Wang
  et~al\mbox{.}}{2020}]%
        {wang2020global}
\bibfield{author}{\bibinfo{person}{Wentao Wang}, \bibinfo{person}{Suhang Wang},
  \bibinfo{person}{Wenqi Fan}, \bibinfo{person}{Zitao Liu}, {and}
  \bibinfo{person}{Jiliang Tang}.} \bibinfo{year}{2020}\natexlab{}.
\newblock \showarticletitle{Global-and-local aware data generation for the
  class imbalance problem}. In \bibinfo{booktitle}{\emph{SDM}}.
\newblock


\bibitem[\protect\citeauthoryear{Wang, Han, Chang, Zha, Braga-Neto, and
  Hu}{Wang et~al\mbox{.}}{2022}]%
        {wang2022auto}
\bibfield{author}{\bibinfo{person}{Yicheng Wang}, \bibinfo{person}{Xiaotian
  Han}, \bibinfo{person}{Chia-Yuan Chang}, \bibinfo{person}{Daochen Zha},
  \bibinfo{person}{Ulisses Braga-Neto}, {and} \bibinfo{person}{Xia Hu}.}
  \bibinfo{year}{2022}\natexlab{}.
\newblock \showarticletitle{Auto-PINN: Understanding and Optimizing
  Physics-Informed Neural Architecture}.
\newblock \bibinfo{journal}{\emph{arXiv preprint arXiv:2205.13748}}
  (\bibinfo{year}{2022}).
\newblock


\bibitem[\protect\citeauthoryear{Wikipedia}{Wikipedia}{2022}]%
        {wiki:Wilcoxon_signed-rank_test}
\bibfield{author}{\bibinfo{person}{Wikipedia}.}
  \bibinfo{year}{2022}\natexlab{}.
\newblock \bibinfo{title}{{Wilcoxon signed-rank test} --- {W}ikipedia{,} The
  Free Encyclopedia}.
\newblock
  \bibinfo{howpublished}{\url{http://en.wikipedia.org/w/index.php?title=Wilcoxon\%20signed-rank\%20test&oldid=1084875027}}.
\newblock
\newblock
\shownote{[Online; accessed 19-May-2022].}


\bibitem[\protect\citeauthoryear{Xu, Skoularidou, Cuesta-Infante, and
  Veeramachaneni}{Xu et~al\mbox{.}}{2019}]%
        {xu2019modeling}
\bibfield{author}{\bibinfo{person}{Lei Xu}, \bibinfo{person}{Maria
  Skoularidou}, \bibinfo{person}{Alfredo Cuesta-Infante}, {and}
  \bibinfo{person}{Kalyan Veeramachaneni}.} \bibinfo{year}{2019}\natexlab{}.
\newblock \showarticletitle{Modeling Tabular data using Conditional GAN}. In
  \bibinfo{booktitle}{\emph{NeurIPS}}.
\newblock


\bibitem[\protect\citeauthoryear{Yang, Akimoto, Kim, and Udell}{Yang
  et~al\mbox{.}}{2019}]%
        {yang2019oboe}
\bibfield{author}{\bibinfo{person}{Chengrun Yang}, \bibinfo{person}{Yuji
  Akimoto}, \bibinfo{person}{Dae~Won Kim}, {and} \bibinfo{person}{Madeleine
  Udell}.} \bibinfo{year}{2019}\natexlab{}.
\newblock \showarticletitle{OBOE: Collaborative filtering for AutoML model
  selection}. In \bibinfo{booktitle}{\emph{KDD}}.
\newblock


\bibitem[\protect\citeauthoryear{Yang, Fan, Wu, and Udell}{Yang
  et~al\mbox{.}}{2020}]%
        {yang2020automl}
\bibfield{author}{\bibinfo{person}{Chengrun Yang}, \bibinfo{person}{Jicong
  Fan}, \bibinfo{person}{Ziyang Wu}, {and} \bibinfo{person}{Madeleine Udell}.}
  \bibinfo{year}{2020}\natexlab{}.
\newblock \showarticletitle{Automl pipeline selection: Efficiently navigating
  the combinatorial space}. In \bibinfo{booktitle}{\emph{KDD}}.
\newblock


\bibitem[\protect\citeauthoryear{Yen and Lee}{Yen and Lee}{2006}]%
        {yen2006under}
\bibfield{author}{\bibinfo{person}{Show-Jane Yen} {and}
  \bibinfo{person}{Yue-Shi Lee}.} \bibinfo{year}{2006}\natexlab{}.
\newblock \showarticletitle{Under-sampling approaches for improving prediction
  of the minority class in an imbalanced dataset}.
\newblock In \bibinfo{booktitle}{\emph{Intelligent Control and Automation}}.
  \bibinfo{publisher}{Springer}, \bibinfo{pages}{731--740}.
\newblock


\bibitem[\protect\citeauthoryear{Yu and Zhu}{Yu and Zhu}{2020}]%
        {yu2020hyper}
\bibfield{author}{\bibinfo{person}{Tong Yu} {and} \bibinfo{person}{Hong Zhu}.}
  \bibinfo{year}{2020}\natexlab{}.
\newblock \showarticletitle{Hyper-parameter optimization: A review of
  algorithms and applications}.
\newblock \bibinfo{journal}{\emph{arXiv preprint arXiv:2003.05689}}
  (\bibinfo{year}{2020}).
\newblock


\bibitem[\protect\citeauthoryear{Zha, Feng, Bhushanam, Choudhary, Nie, Tian,
  Chae, Ma, Kejariwal, and Hu}{Zha et~al\mbox{.}}{2022a}]%
        {zha2022autoshard}
\bibfield{author}{\bibinfo{person}{Daochen Zha}, \bibinfo{person}{Louis Feng},
  \bibinfo{person}{Bhargav Bhushanam}, \bibinfo{person}{Dhruv Choudhary},
  \bibinfo{person}{Jade Nie}, \bibinfo{person}{Yuandong Tian},
  \bibinfo{person}{Jay Chae}, \bibinfo{person}{Yinbin Ma},
  \bibinfo{person}{Arun Kejariwal}, {and} \bibinfo{person}{Xia Hu}.}
  \bibinfo{year}{2022}\natexlab{a}.
\newblock \showarticletitle{AutoShard: Automated Embedding Table Sharding for
  Recommender Systems}. In \bibinfo{booktitle}{\emph{KDD}}.
\newblock


\bibitem[\protect\citeauthoryear{Zha, Lai, Huang, Cao, Reddy, Vargas, Nguyen,
  Wei, Guo, and Hu}{Zha et~al\mbox{.}}{2021a}]%
        {zha2021rlcard}
\bibfield{author}{\bibinfo{person}{Daochen Zha}, \bibinfo{person}{Kwei-Herng
  Lai}, \bibinfo{person}{Songyi Huang}, \bibinfo{person}{Yuanpu Cao},
  \bibinfo{person}{Keerthana Reddy}, \bibinfo{person}{Juan Vargas},
  \bibinfo{person}{Alex Nguyen}, \bibinfo{person}{Ruzhe Wei},
  \bibinfo{person}{Junyu Guo}, {and} \bibinfo{person}{Xia Hu}.}
  \bibinfo{year}{2021}\natexlab{a}.
\newblock \showarticletitle{RLCard: a platform for reinforcement learning in
  card games}. In \bibinfo{booktitle}{\emph{IJCAI}}.
\newblock


\bibitem[\protect\citeauthoryear{Zha, Lai, Wan, and Hu}{Zha
  et~al\mbox{.}}{2020}]%
        {zha2020meta}
\bibfield{author}{\bibinfo{person}{Daochen Zha}, \bibinfo{person}{Kwei-Herng
  Lai}, \bibinfo{person}{Mingyang Wan}, {and} \bibinfo{person}{Xia Hu}.}
  \bibinfo{year}{2020}\natexlab{}.
\newblock \showarticletitle{Meta-AAD: Active anomaly detection with deep
  reinforcement learning}. In \bibinfo{booktitle}{\emph{ICDM}}.
\newblock


\bibitem[\protect\citeauthoryear{Zha, Lai, Zhou, and Hu}{Zha
  et~al\mbox{.}}{2019}]%
        {zha2019experience}
\bibfield{author}{\bibinfo{person}{Daochen Zha}, \bibinfo{person}{Kwei-Herng
  Lai}, \bibinfo{person}{Kaixiong Zhou}, {and} \bibinfo{person}{Xia Hu}.}
  \bibinfo{year}{2019}\natexlab{}.
\newblock \showarticletitle{Experience Replay Optimization}. In
  \bibinfo{booktitle}{\emph{IJCAI}}.
\newblock


\bibitem[\protect\citeauthoryear{Zha, Lai, Zhou, and Hu}{Zha
  et~al\mbox{.}}{2021b}]%
        {zha2021simplifying}
\bibfield{author}{\bibinfo{person}{Daochen Zha}, \bibinfo{person}{Kwei-Herng
  Lai}, \bibinfo{person}{Kaixiong Zhou}, {and} \bibinfo{person}{Xia Hu}.}
  \bibinfo{year}{2021}\natexlab{b}.
\newblock \showarticletitle{Simplifying deep reinforcement learning via
  self-supervision}.
\newblock \bibinfo{journal}{\emph{arXiv preprint arXiv:2106.05526}}
  (\bibinfo{year}{2021}).
\newblock


\bibitem[\protect\citeauthoryear{Zha, Ma, Yuan, Hu, and Liu}{Zha
  et~al\mbox{.}}{2021c}]%
        {zha2021rank}
\bibfield{author}{\bibinfo{person}{Daochen Zha}, \bibinfo{person}{Wenye Ma},
  \bibinfo{person}{Lei Yuan}, \bibinfo{person}{Xia Hu}, {and}
  \bibinfo{person}{Ji Liu}.} \bibinfo{year}{2021}\natexlab{c}.
\newblock \showarticletitle{Rank the Episodes: A Simple Approach for
  Exploration in Procedurally-Generated Environments}. In
  \bibinfo{booktitle}{\emph{ICLR}}.
\newblock


\bibitem[\protect\citeauthoryear{Zha, Pervaiz~Bhat, Chen, Wang, Ding, Jain,
  Qazim~Bhat, Lai, Chen, et~al\mbox{.}}{Zha et~al\mbox{.}}{2022b}]%
        {zha2021autovideo}
\bibfield{author}{\bibinfo{person}{Daochen Zha}, \bibinfo{person}{Zaid
  Pervaiz~Bhat}, \bibinfo{person}{Yi-Wei Chen}, \bibinfo{person}{Yicheng Wang},
  \bibinfo{person}{Sirui Ding}, \bibinfo{person}{Anmoll~Kumar Jain},
  \bibinfo{person}{Mohammad Qazim~Bhat}, \bibinfo{person}{Kwei-Herng Lai},
  \bibinfo{person}{Jiaben Chen}, {et~al\mbox{.}}}
  \bibinfo{year}{2022}\natexlab{b}.
\newblock \showarticletitle{AutoVideo: An Automated Video Action Recognition
  System}. In \bibinfo{booktitle}{\emph{IJCAI}}.
\newblock


\bibitem[\protect\citeauthoryear{Zha, Xie, Ma, Zhang, Lian, Hu, and Liu}{Zha
  et~al\mbox{.}}{2021d}]%
        {zha2021douzero}
\bibfield{author}{\bibinfo{person}{Daochen Zha}, \bibinfo{person}{Jingru Xie},
  \bibinfo{person}{Wenye Ma}, \bibinfo{person}{Sheng Zhang},
  \bibinfo{person}{Xiangru Lian}, \bibinfo{person}{Xia Hu}, {and}
  \bibinfo{person}{Ji Liu}.} \bibinfo{year}{2021}\natexlab{d}.
\newblock \showarticletitle{DouZero: Mastering DouDizhu with Self-Play Deep
  Reinforcement Learning}. In \bibinfo{booktitle}{\emph{ICML}}.
\newblock


\bibitem[\protect\citeauthoryear{Zhao, Zhang, and Wang}{Zhao
  et~al\mbox{.}}{2021c}]%
        {zhao2021graphsmote}
\bibfield{author}{\bibinfo{person}{Tianxiang Zhao}, \bibinfo{person}{Xiang
  Zhang}, {and} \bibinfo{person}{Suhang Wang}.}
  \bibinfo{year}{2021}\natexlab{c}.
\newblock \showarticletitle{Graphsmote: Imbalanced node classification on
  graphs with graph neural networks}. In \bibinfo{booktitle}{\emph{WSDM}}.
\newblock


\bibitem[\protect\citeauthoryear{Zhao, Liu, Fan, Liu, Tang, Wang, Chen, Zheng,
  Liu, and Yang}{Zhao et~al\mbox{.}}{2021a}]%
        {zhaok2021autoemb}
\bibfield{author}{\bibinfo{person}{Xiangyu Zhao}, \bibinfo{person}{Haochen
  Liu}, \bibinfo{person}{Wenqi Fan}, \bibinfo{person}{Hui Liu},
  \bibinfo{person}{Jiliang Tang}, \bibinfo{person}{Chong Wang},
  \bibinfo{person}{Ming Chen}, \bibinfo{person}{Xudong Zheng},
  \bibinfo{person}{Xiaobing Liu}, {and} \bibinfo{person}{Xiwang Yang}.}
  \bibinfo{year}{2021}\natexlab{a}.
\newblock \showarticletitle{Autoemb: Automated embedding dimensionality search
  in streaming recommendations}. In \bibinfo{booktitle}{\emph{ICDM}}.
\newblock


\bibitem[\protect\citeauthoryear{Zhao, Rossi, and Akoglu}{Zhao
  et~al\mbox{.}}{2021b}]%
        {zhao2021automatic}
\bibfield{author}{\bibinfo{person}{Yue Zhao}, \bibinfo{person}{Ryan Rossi},
  {and} \bibinfo{person}{Leman Akoglu}.} \bibinfo{year}{2021}\natexlab{b}.
\newblock \showarticletitle{Automatic unsupervised outlier model selection}.
\newblock \bibinfo{journal}{\emph{NeurIPS}} (\bibinfo{year}{2021}).
\newblock


\end{thebibliography}

\end{document}